%% file: paper.tex
\documentclass{article}

\usepackage{microtype}
\usepackage{graphicx}
\usepackage{subfigure}
\usepackage{booktabs} % for professional tables

\usepackage{lipsum}
\usepackage{amsmath}
\usepackage{colortbl}
\usepackage{multirow}
\usepackage[table,xcdraw]{xcolor}
\usepackage{subfiles}
\usepackage{threeparttable}

\usepackage{chngcntr}
\usepackage{bibentry}
\usepackage{rotating}
\usepackage{diagbox}

\usepackage{hyperref}

\usepackage[accepted]{icml2020}

\icmltitlerunning{Rethinking Empirical Evaluation of Adversarial Robustness Using First-Order Attack Methods}

\begin{document}

\twocolumn[
\icmltitle{Rethinking Empirical Evaluation of Adversarial Robustness \\ Using First-Order Attack Methods}



\icmlsetsymbol{equal}{*}

\begin{icmlauthorlist}
\icmlauthor{Kyungmi Lee}{to}
\icmlauthor{Anantha P. Chandrakasan}{to}
\end{icmlauthorlist}

\icmlaffiliation{to}{Department of Electrical Engineering and Computer Science, Massachusetts Institute of Technology, Cambridge, MA, USA}

\icmlcorrespondingauthor{Kyungmi Lee}{kyungmi@mit.edu}

\icmlkeywords{Adversarial Robustness, Gradient Masking, First-Order Method}

\vskip 0.3in
]



\printAffiliationsAndNotice{}  % leave blank if no need to mention equal contribution

\begin{abstract}
\subfile{writing/0-Abstract.tex}
%\lipsum[1]
\end{abstract}

\section{Introduction}
\subfile{writing/1-Intro.tex}
%\lipsum[2-6]

\section{Preliminaries}
\subsection{Notations and definitions}
\subfile{writing/2-1.tex}
\subsection{Experimental setup}
\subfile{writing/2-2.tex}

\section{Analysis on failure modes of attacks}
\subfile{writing/3-Header.tex}
\subsection{Zero loss}
\subfile{writing/3-1.tex}
\subsection{Innate non-differentiability}
\subfile{writing/3-2.tex}
\subsection{Require more iterations}
\subfile{writing/3-3.tex}

\section{Case study}
\subfile{writing/4-Header.tex}
\subsection{Model capacity}
\subsubsection{Training models with different width}
\subfile{writing/4-1-1.tex}
\subsubsection{Weight pruning}
\subfile{writing/4-1-2.tex}
\subsection{Regularization}
\subfile{writing/4-2.tex}

\section{Transferability of compensated attacks}
\subfile{writing/5-blackbox.tex}

\section{Comparison with verified lower bounds}
\subfile{writing/6-verification.tex}

\section{Related work}
\subfile{writing/7-related.tex}

\section{Conclusion}
\subfile{writing/8-Conclusion.tex}


\bibliography{paper}
\bibliographystyle{icml2020}

\onecolumn
\newpage
\appendix
\counterwithin{figure}{section}
\counterwithin{table}{section}
\numberwithin{equation}{section}
\section{Details of experimental setups}
\subsection{Architectures}
\subfile{app-writing/a-1.tex}
\subsection{Training condition}
\subfile{app-writing/a-2.tex}
\subsection{Equations for attack methods}
\subfile{app-writing/a-3.tex}

\section{Additional analysis on failure of attacks}
\subsection{Zero loss}
\subsubsection{Omitted analysis}
\subfile{app-writing/b-1-1.tex}
\subsubsection{Comparison of compensation methods}
\subfile{app-writing/b-1-2.tex}
\subsubsection{Comparison of different models}
\subfile{app-writing/b-1-3.tex}

\subsection{Innate non-differentiability}
\subfile{app-writing/b-2.tex}

\subsection{Require more iterations}
\subfile{app-writing/b-3.tex}

\section{Additional experiments on case studies}
\subfile{app-writing/c.tex}
\subsection{Training models with different width}
\subfile{app-writing/c-1.tex}
\subsection{Weight pruning}
\subfile{app-writing/c-2.tex}
\subsection{Regularization}
\subfile{app-writing/c-3.tex}
\subsubsection{Details on regularization methods}
\subfile{app-writing/c-3-1.tex}
\subsubsection{Additional experiments}
\subfile{app-writing/c-3-2.tex}

\section{Additional experiments on transferability}
\subfile{app-writing/d.tex}

\section{Experimental details on verified lower bounds}
\subfile{app-writing/part-e.tex}

\section{C\&W attack and zero loss}
\subfile{app-writing/f.tex}

\end{document}

%% file: writing/0-Abstract.tex
We identify three common cases that lead to overestimation of adversarial accuracy against bounded first-order attack methods, which is popularly used as a proxy for adversarial robustness in empirical studies. For each case, we propose compensation methods that either address sources of inaccurate gradient computation, such as numerical instability near zero and non-differentiability, or reduce the total number of back-propagations for iterative attacks by approximating second-order information. These compensation methods can be combined with existing attack methods for a more precise empirical evaluation metric. We illustrate the impact of these three cases with examples of practical interest, such as benchmarking model capacity and regularization techniques for robustness. Overall, our work shows that overestimated adversarial accuracy that is not indicative of robustness is prevalent even for conventionally trained deep neural networks, and highlights cautions of using empirical evaluation without guaranteed bounds. 

%% file: writing/1-Intro.tex
Robustness against adversarial examples \cite{szegedy2013intriguing, Biggio_2018} is becoming an important factor for designing deep neural networks, resulting in increased interest in benchmarking architectures \cite{Su2018IsRT} and regularization techniques \cite{madry2017towards, jakubovitz2018improving} for robustness. An essential but challenging methodology for understanding adversarial robustness is precise evaluation of adversarial robustness. For a bounded adversarial example (i.e., within an $\epsilon$-ball in $L_p$ norm from an unperturbed input sample $x$), adversarial robustness boils down to the existence of a perturbation $r$ ($\|r\|_p < \epsilon$) such that a deep neural network's predictions on $x$ and $x+r$ are different, which is difficult to (dis)prove for a high-dimensional input sample. Consequently, empirically testing the prediction using $r$ generated by bounded first-order attack methods such as Fast Gradient Sign Method \cite{goodfellow2014explaining} and Projected Gradient Descent \cite{madry2017towards}, which are algorithms that generate $r$ efficiently with few back-propagations, became a popular approach. Then, accuracy against samples generated by those attack methods is taken as a proxy for adversarial robustness for the entire dataset (e.g., test set). 

Nevertheless, this approach only yields an upper bound of robustness; that is, failure of attack methods to find adversarial examples might not imply true robustness. Notably, gradient masking \cite{papernot_gradmask, tramr2017ensemble, athalye2018obfuscated} inflates adversarial accuracy by inducing gradients used by attack methods to be inaccurate. Therefore, it is important to understand when failure of attack methods is \emph{not} indicative of robustness in order to obtain more precise metric for empirical evaluation of adversarial robustness.

In this work, we identify three common cases in which bounded first-order attack methods are unsuccessful due to superficial reasons that do not indicate robustness: 1) cross-entropy loss close to zero resulting in inaccurate computation of gradients, 2) gradient shattering \cite{athalye2018obfuscated} induced by non-differentiable activation and pooling operations, and 3) certain training conditions inducing deep neural networks to be less amenable to first-order approximation, increasing the number of iterations for iterative attacks to successfully find adversarial examples. We observe these phenomena are prevalent in various conventionally trained deep neural networks across different architectures and datasets, not only confined to specific defenses intentionally designed to cause gradient masking. For each case, we propose compensation methods to address the cause (Section 3).

We demonstrate the impact of these phenomena using case studies on model capacity and regularization techniques (Section 4). We further analyze transferability of compensation methods for black-box scenarios (Section 5), and whether these phenomena can explain the gap between empirical adversarial accuracy and verified lower bounds of robustness \cite{wong2017provable, tjeng2018evaluating} (Section 6). We conclude this paper by linking our finding back to related work (Section 7), and with a short remark on future work (Section 8). 

%% file: writing/2-1.tex
Throughout this paper, we use $x$ to denote an unperturbed (clean) input  and $f(\cdot)$ to represent a neural network. Also, we denote output pre-softmax logits of a neural network given $x$ as $z=f(x)$. We only consider the classification task, and use cross-entropy loss for training and generating adversarial samples. Naturally, the predicted label is $y=\text{arg}\underset{i}{\text{max}}z_i$ where $z_i$ represents the value of $i$th logit. Given logits $z$ and the ground truth label $t$, we express the loss as $l(z, t)$. Gradients of the loss with respect to $x$ are denoted as $g:= \frac{\partial l(z, t)}{\partial x} = \frac{\partial l(f(x), t)}{\partial x}$. 

We consider adversarial robustness within an $\epsilon$-ball defined for $L_p$ norm around an input sample $x$ that is correctly predicted by a neural network $f$. That is, if there exists a perturbation $r$ such that $\|r\|_p < \epsilon$ and $f(x+r)\ne f(x)$, we claim that $f$ is \emph{not} robust for $x$.

%% file: writing/2-2.tex
\textbf{Dataset} We use CIFAR-10 \cite{krizhevsky2009learning}, SVHN \cite{netzer2011reading}, and TinyImageNet (a down-sampled dataset from \citeauthor{ILSVRC15} (\citeyear{ILSVRC15})) to analyze and benchmark adversarial accuracy. The images are normalized to the range $[0, 1]$ for both training and testing, and further pre-processing includes random crop and flips during training. We randomly split 10\% of training samples for validation purpose. For Section 6, we additionally use MNIST \cite{lecun1998gradient} and follow pre-processing of \citeauthor{wong2017provable} (\citeyear{wong2017provable}) for MNIST and CIFAR-10.

\textbf{Neural network architectures} We examine various design choices, including different model capacity, usage of batch normalization \cite{ioffe2015batch}, and residual connections \cite{He_2016}. For CIFAR-10 and SVHN, we consider a Simple model with 4 convolutional layers and 2 fully connected layers, a Simple-BN model that has a batch normalization layer following each convolutional layer in a Simple model, and a WideResNet (WRN) \cite{zagoruyko2016wide} with depth of 28 layers. For TinyImageNet, we use a VGG-11 \cite{simonyan2014deep} with and without batch normalization, and a WRN with depth of 50 layers. Details of architectures and their training hyperparameters are described in Appendix A. 

\textbf{Adversarial attack methods} We examine bounded first-order attack methods that are popularly used to study adversarial robustness for $L_2$ and $L_\infty$ norms, for their untargeted adversarial examples, primarily under full white-box setting. In particular, we consider Fast Gradient Sign Method (FGSM) \cite{goodfellow2014explaining}, Random-FGSM (R-FGSM) \cite{tramr2017ensemble}, and Projected Gradient Descent (PGD) \cite{madry2017towards}. FGSM computes a perturbation $r$ as $r=\epsilon\cdot\texttt{sign}(g)$ for $L_\infty$ norm using a single back-propagation to obtain $g$. R-FGSM modifies FGSM by adding a random perturbation before computing gradients. PGD uses iterative update to compute a perturbation after random initialization, and each iteration evaluates gradients as in FGSM. For $L_2$ norm, \texttt{sign} is replaced with dividing by $\|g\|_2$ to produce a unit vector.

\textbf{Implementation}\footnote{The source code for this paper is available at \url{https://github.com/kyungmi-lee/eval-adv-robustness}.} All experiments are implemented with PyTorch \cite{paszke2017automatic}. We use AdverTorch \cite{ding2019advertorch} as a framework to implement attack methods. 

%% file: writing/3-Header.tex
In this section, we analyze when bounded first-order attack methods fail to find adversarial examples, and identify three cases in which such failure does not indicate robustness. We provide compensation methods for each case to improve the evaluation metric. 

%% file: writing/3-1.tex
First-order attack methods use gradients to compute the perturbation direction. However, when the value of loss function becomes zero, gradients are naturally zero and do not provide meaningful information about the perturbation direction. Similar phenomenon occurs when the value of loss function is not exactly zero, but very small so that limited numerical precision and stability contaminate gradients to be no longer useful. Here we analyze this (near) zero loss phenomenon and propose simple methods to compensate for this phenomenon.

\subfile{writing/table-3-1.tex}
\textbf{Analysis} Cross-entropy loss gets small when pre-softmax logits $z$ have large ``margin", the gap between logits corresponding to the most likely and the second most likely labels, and often becomes zero due to limited numerical precision. However, logit margins can be simply inflated by weight matrices with large values, for instance when no regularization that can penalize large weights is applied. To illustrate, if a matrix $W$ for a linear equation $u=Wv$ gets multiplied by a constant $c > 1$, the difference between $i$th and $j$th element of $u$ will also be inflated by $c$. Taking exponential on $z$ with large margin to compute loss can easily result in near zero loss.

From an experiment using a Simple model trained without explicit regularization on CIFAR-10, we also observe a symptom of gradient masking, specifically a black-box attack being stronger than a white-box attack \cite{athalye2018obfuscated} (full detail in Appendix B.1). While this implies that adversarial accuracy of the model could have been overestimated, we also find many samples on which an attack (e.g., FGSM) fails have (near) zero loss ($<10^{-8}$) induced by large logit margin. Overall, we cannot conclude that failure of attacks to find adversarial examples in case of (near) zero loss is indicative of robustness. 

\textbf{Compensation} A straightforward way to account for this (near) zero loss phenomenon is to ensure that the value of loss is sufficiently large so that gradients can be computed accurately. First, we consider rescaling pre-softmax logits $z$ by dividing with a fixed constant $T$; that is, we compute gradients for loss on rescaled logits, $l(\frac{f(x)}{T}, t)$ where $t$ is ground truth labels. When $T > 1$, absolute value of logits decrease, leading to larger cross-entropy loss. Alternatively, we consider changing $t$ from ground truth labels to other classes when computing loss, which essentially gives same expression as targeted attacks. Since loss with respect to ground truth labels is small, changing target class will increase loss. Note that still our interest is to find \emph{untargeted} adversarial examples although we change loss function to be targeted to ensure large value. After increasing the value of loss using either approach, we apply first-order attack methods as usual. 

We find that changing target labels to be the second most likely classes generally gives the best compensation method for white-box attacks compared to rescaling logits and other possible target labels (e.g. randomly chosen classes or the least likely classes), increasing success rate of attack methods up to $11\%$ for FGSM and $4.5\%$ for PGD ($\epsilon=\frac{8}{255}$, $L_\infty$) for the above-mentioned Simple model. Quantitative comparison of these methods is presented in Appendix B.1.

\textbf{Impact} We benchmark how this phenomenon affects evaluation against first-order attack methods by comparing adversarial accuracy against baseline attacks (i.e., vanilla FGSM, R-FGSM, and PGD) and attacks with the compensation method using the second most likely classes as targets for computing loss. We apply the compensation method to samples on which baseline attacks fail and report resulting accuracy (Table \ref{table:3-1}, Column `Zero loss'). 

Gradient masking induced by zero loss was previously observed by \citeauthor{Carlini_2017} (\citeyear{Carlini_2017}) for their analysis of defensive distillation \cite{Papernot_2016}, which used temperature softmax (same as rescaling logits) during the distillation step. However, we find that this phenomenon is not confined to specific defenses that deliberately caused gradient masking; we can see prevalence of this phenomenon among many conventionally trained models as shown in Table \ref{table:3-1}. 

We can expect that this phenomenon will affect models trained without explicit regularization that have weight matrices with large magnitudes more severely compared to models trained with regularizations that penalize large weights such as weight decay. For a Simple model trained on CIFAR-10, we verify that compensating for this phenomenon decreases adversarial accuracy by $4.48\%$ and $0.19\%$ for FGSM and PGD ($\epsilon=\frac{4}{255}$, $L_\infty$) when weight decay is applied, compared to $8.71\%$ and $0.87\%$ when no regularization is used.

%% file: writing/table-3-1.tex
% \usepackage{booktabs}
% \usepackage{multirow}
\begin{table*}[t]
\centering
\caption{Accuracy of neural networks against first-order attack methods in the order of FGSM/R-FGSM/PGD for $\epsilon=\frac{4}{255} \text{ in } L_\infty$ norm. We apply compensation methods for zero loss and innate non-differentiability discussed in Sections 3.1 and 3.2. As compensation methods are applied in cascading manner (i.e., samples on which baseline attacks fail are subjected to compensation methods), we set the number of random starts for baseline R-FGSM and PGD to be same as the total number of evaluations compensation methods use for fair comparison. All models are trained without explicit regularization on the specified dataset.}
\small
\vspace{0.5em}
\begin{tabular}{@{}ccccccc@{}}
\toprule
\multirow{2}{*}{Dataset}      & \multirow{2}{*}{Architecture} & \multicolumn{5}{c}{Accuracy (\%)}                                                                  \\ \cmidrule(l){3-7} 
                              &                               & Clean & Attack Baseline      & Zero loss            & Non-differentiability & Both                 \\ \midrule
\multirow{3}{*}{CIFAR-10}     & Simple                        & 84.75 & 19.50 / 29.81 / 2.55 & 10.79 / 29.11 / 1.68 & 18.41 / 29.17 / 2.54  & 8.74 / 28.31 / 1.67  \\
                              & Simple-BN                     & 87.09 & 28.66 / 28.39 / 6.26 & 9.93 / 19.03 / 0.10  & 26.92 / 28.05 / 6.11  & 6.89 / 17.93 / 0.07  \\
                              & WRN 28                 & 91.65 & 21.90 / 20.79 / 0.02 & 11.34 / 13.46 / 0    & 15.87 / 19.76 / 0.02  & 5.94 / 11.25 / 0     \\ \midrule
\multirow{2}{*}{SVHN}         & Simple-BN                     & 94.29 & 28.60 / 43.36 / 2.80 & 23.81 / 42.21 / 2.59 & 24.35 / 42.60 / 2.60  & 19.10 / 41.31 / 2.38 \\
                              & WRN 28                 & 95.42 & 49.74 / 58.22 / 4.06 & 45.46 / 57.04 / 3.73 & 39.69 / 56.70 / 3.79  & 34.30 / 55.05 / 3.69 \\ \midrule
\multirow{3}{*}{TinyImageNet} & VGG 11                        & 50.32 & 11.32 / 20.16 / 7.94 & 7.44 / 16.56 / 4.10  & 11.30 / 20.62 / 7.98  & 7.44 / 16.86 / 4.22  \\
                              & VGG-BN 11                     & 50.72 & 4.16 / 11.68 / 0.64  & 3.22 / 10.70 / 0.48  & 3.82 / 11.80 / 0.68   & 3.06 / 11.00 / 0.48  \\
                              & WRN 50                 & 57.24 & 19.78 / 23.40 / 2.02 & 3.36 / 9.86 / 0.40   & 18.24 / 23.48 / 1.58  & 2.88 / 10.02 / 0.36  \\ \bottomrule
\end{tabular}
\label{table:3-1}
\end{table*}

%% file: writing/3-2.tex
Non-differentiability of functions that are in the computation graph for back-propagation causes gradient shattering, a type of gradient masking identified by \citeauthor{athalye2018obfuscated} (\citeyear{athalye2018obfuscated}). In this section, we analyze how innate non-differentiability induced by popularly used Rectified Linear Unit (ReLU) and max pooling subtly affects attack methods, and how to compensate for this phenomenon using Backward Pass Differentiable Attack (BPDA) \cite{athalye2018obfuscated}.

\textbf{Analysis} A layer using ReLU as an activation function passes gradients only through non-negative valued neurons, as negative valued neurons are set to be zero after ReLU. However, adding perturbations $r$ can ``switch" some of negative valued neurons, which did not contribute to gradients originally, to take non-negative values (or vice versa) during forward-propagation. In such cases, gradients are no longer accurate as the effective neurons contributing to the final prediction are changed. A similar problem exists for max pooling when perturbations change the maximum valued neurons in each pooling window.

We analyze how often this switching happens for ReLU and max pooling when perturbations are added to inputs. On a Simple model trained without explicit regularization on CIFAR-10, we observe that more neurons switch for larger perturbation size $\epsilon$, and the fraction of neurons that switch can be significant, especially for max pooling ($7.73\%$ of ReLU neurons and $20.60\%$ of max pooling neurons switch for a FGSM attack with $\epsilon=\frac{8}{255}$ in $L_\infty$ norm). 

\textbf{Compensation} BPDA provides a method to approximate gradients for non-differentiable functions, by substituting such functions with similar but differentiable functions during back-propagation. Although BPDA was originally used to break defenses that relied on non-differentiability, we use BPDA to smoothly approximate behaviors of ReLU and max pooling around their non-differentiable points. Substituting ReLU with softplus and max pooling with $L_p$ norm pooling with $p=5$ can be used as a default setting for BPDA especially when compensating for both this and the zero loss phenomena. Detailed comparison of substitute functions are shown in Appendix B.2.

\textbf{Impact} Similar to the zero loss phenomenon, we benchmark how innate non-differentiability affects adversarial accuracy using softplus and $p=5 \text{ or } 10$ as the compensation method (Table \ref{table:3-1}, Column `Non-differentiability'). We observe that non-differentiability generally gives subtle difference ($\sim1\%$) especially for R-FGSM and PGD. Nevertheless, there are models significantly affected by non-differentiability, such as a WRN 28 trained on SVHN, which shows more than $10\%$ difference for FGSM. Also, FGSM seems to be more affected than other attack methods, which can be explained by smaller effective step size of R-FGSM and PGD, since smaller $\epsilon$ results in less frequent switching of neurons. For example, PGD uses small step size ($<\epsilon$) for each iteration, and R-FGSM typically takes a step size of $\frac{\epsilon}{2}$ instead of $\epsilon$ due to added random perturbation.

We also report adversarial accuracy when attacks are compensated for both zero loss and non-differentiability (Table \ref{table:3-1}, Column `Both'). Compensating for both phenomena additionally accounts for up to $5\%$ (e.g., WRN 28 models trained on CIFAR-10 and SVHN evaluated against FGSM) compared to only compensating for one of them. 

%% file: writing/3-3.tex
We observe that certain training conditions increase the number of iterations required to find adversarial examples for iterative attack methods, resulting in increased adversarial accuracy when evaluated against attacks using small fixed number of iterations. We propose to incorporate second-order information to reduce the total number of back-propagations when this phenomenon occurs. 

\subfile{writing/figure-3-3.tex}
\textbf{Analysis} We find that applying weight decay excessively (increasing strength of weight decay as training progresses) for a WRN 28 trained on CIFAR-10 improves adversarial accuracy against PGD ($\epsilon=0.5, L_2$, compensated for zero loss and non-differentiability) using small number of iterations, but eventually does not manifest in a benefit compared to a model trained without explicit regularization for large number of iterations (Figure \ref{fig:3-3}, above). For example, a model with excessive weight decay is evaluated to have $7.11\%$ of adversarial accuracy compared to only $0.12\%$ of a model with no regularization when PGD uses 5 iterations, but it ultimately shows less than $0.1\%$ of adversarial accuracy when the number of iterations is increased to 100. This observation can be concerning when one uses attacks with a small number of iterations, which can be typical when evaluating complex neural networks where back-propagations are computationally expensive, to compare different training conditions and concludes that simply applying weight decay excessively can provide advantage. 

\textbf{Compensation} While using a large number of iterations is an uncomplicated way to prevent overestimation of adversarial accuracy, we investigate methods to reduce the number of back-propagations needed and to better understand why certain conditions require more iterations. A couple of plain observations are that random initialization of PGD affects the success of subsequent first-order iterations, which is the well-known reason for using multiple random starts when attack methods have stochasticity, and that successful initialization (from which first-order iterations find adversarial examples) leads to larger increase of loss and the \emph{size of gradients} (in $L_2$ norm) per iteration. Although the observations themselves can be trivial, they hint that initializing to a point where gradients change rapidly, thus with high curvature, might help subsequent first-order iterations to find adversarial examples easily.

However, exact curvature of loss is second-order information that is computationally expensive and numerically unstable to obtain. Thus we consider a method that can approximate the principal eigenvector (corresponding to the largest eigenvalue) of the Hessian, which provides information on which direction loss changes fastest. We adopt a single step of power iteration (Eq \eqref{eq1}) and finite difference method (Eq \eqref{miyatoeq2}) of \citeauthor{Miyato_2019} (\citeyear{Miyato_2019}), which roughly approximates the principal eigenvector $u$ of Hessian $H$ as:
\begin{gather}
%\begin{align}
    u \leftarrow \frac{H\cdot d}{\|H\cdot d\|_2} \label{eq1}\\
    H\cdot d = \frac{\frac{\partial l(f(x'), t)}{\partial x'}\Bigr|_{x'=x+\delta d} - \frac{\partial l(f(x'), t)}{\partial x'}\Bigr|_{x'=x}}{\delta} \label{miyatoeq2}
%\end{align}
\end{gather}
where $d$ is randomly sampled from $\mathcal{N}(0, I)$ and normalized to be a unit vector. This method uses two additional back-propagations to compute $u$, which we use as a direction for initialization of PGD (PGD + Eigen) instead of a random vector. 

We also examine Quasi-Newton method, specifically BFGS, which approximates the inverse of the Hessian used to compute Newton direction. To simplify computational overhead, we only use a single iteration to update the Hessian, and omit line search and instead use perturbation size $\epsilon$ to obtain initialization (PGD + BFGS). This method also adds two additional back-propagations.

\textbf{Impact} We test the effectiveness of these two methods on the above-mentioned WRN 28 trained with excessive weight decay, by comparing adversarial accuracy against PGD attacks using these methods as initialization, under the same \emph{total} number of back-propagations used for both initialization and first-order iteration (Figure \ref{fig:3-3}, below). We find that both methods provide stronger attack; for example, adversarial accuracy against PGD + Eigen and PGD + BFGS are $4.82\%$ and $3.60\%$ compared to $7.11\%$ of baseline PGD when using only 5 back-propagations (equivalent to 5 iterations for baseline PGD, and 3 first-order iterations for PGD + Eigen and PGD + BFGS). Thus, utilizing approximate second-order information reduces the total number of back-propagations to achieve similar success rate of attacks when this phenomenon occurs.

%% file: writing/figure-3-3.tex
\begin{figure}[t]
    \centering
    \includegraphics[width=0.8\linewidth]{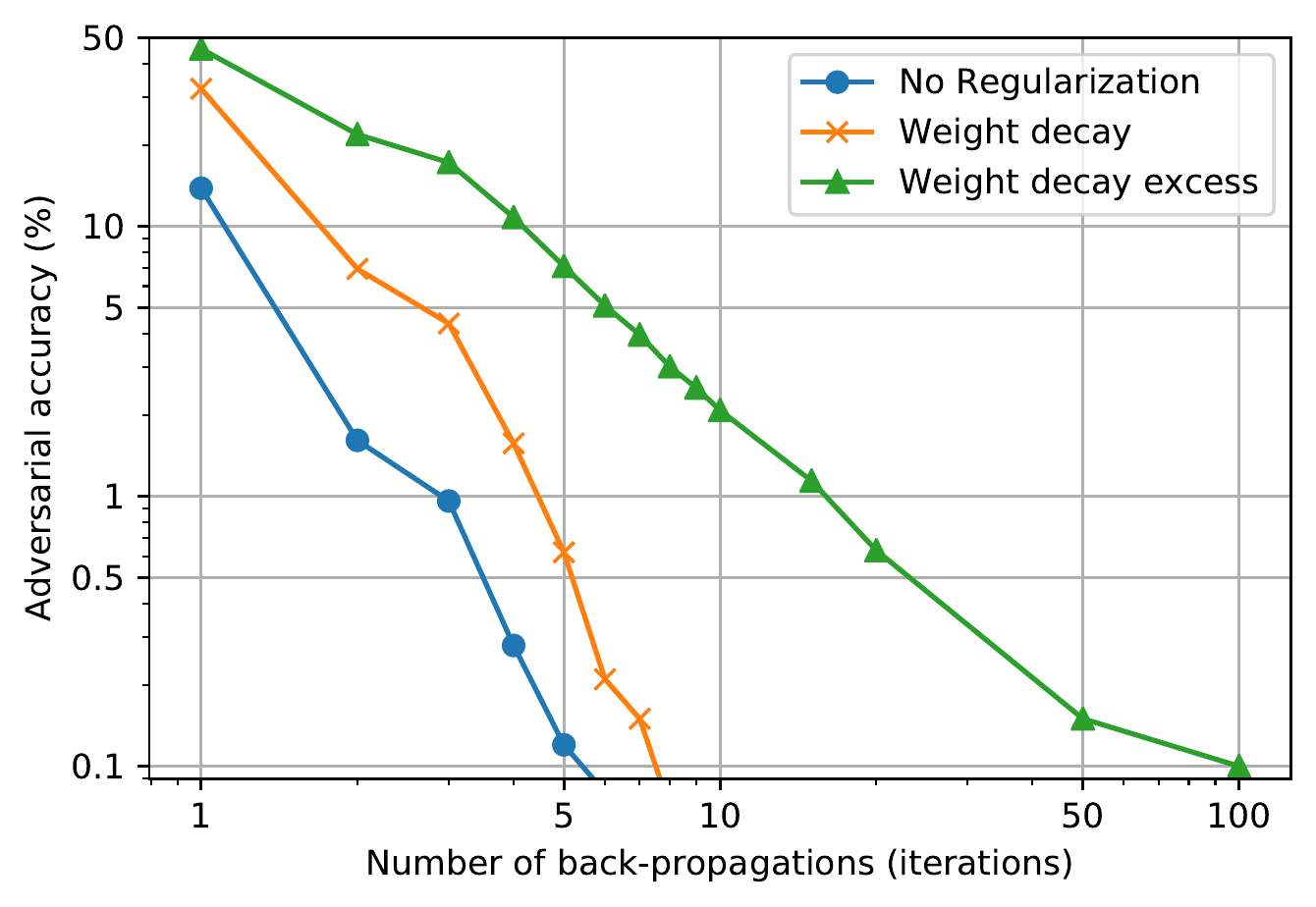}
    \includegraphics[width=0.8\linewidth]{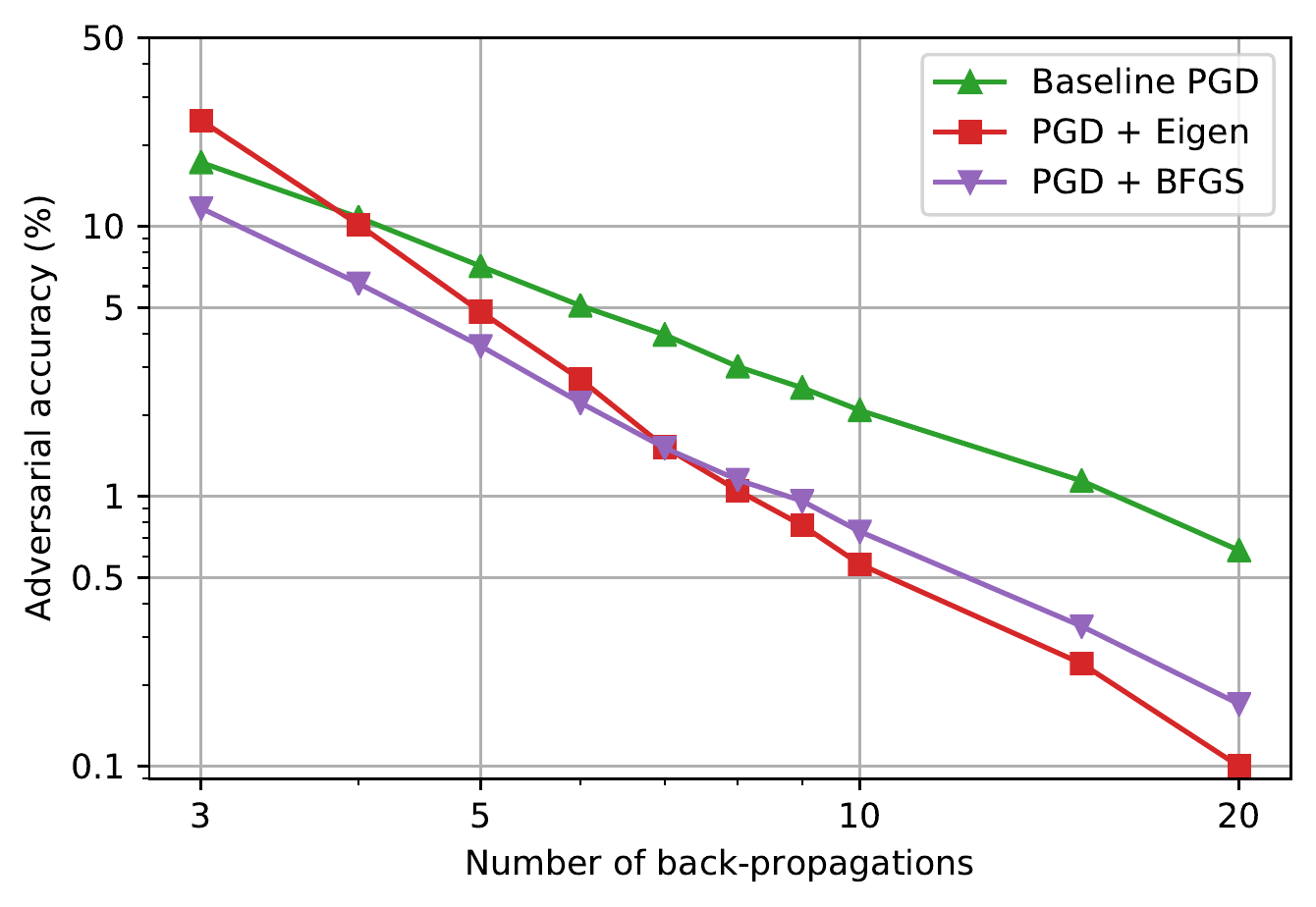}
    \caption{(Above) Adversarial accuracy against a PGD attack ($\epsilon=0.5$, $L_2$, compensated for zero loss and non-differentiability) of WRN 28 models trained on CIFAR-10 with different regularization conditions, as a function of the number of iterations the attack uses. Shown in log-log scale. (Below) Comparison of different compensation methods (Eigen and BFGS) discussed in Section 3.3 on a WRN 28 trained with excessive weight decay (same as in the above figure), under the same number of back-propagations required to find adversarial examples.}
    \label{fig:3-3}
    \vspace{-1em}
\end{figure}

%% file: writing/4-Header.tex
In this section, we investigate how the three phenomena inducing overestimation of adversarial accuracy affect practically important cases, such as benchmarking the trade-off between model capacity and adversarial robustness, and comparison of regularization techniques.

\subfile{writing/figure-4-1.tex}
\subfile{writing/figure-4-2.tex}

%% file: writing/figure-4-1.tex
\begin{figure}[t]
    \centering
    \includegraphics[width=0.8\linewidth]{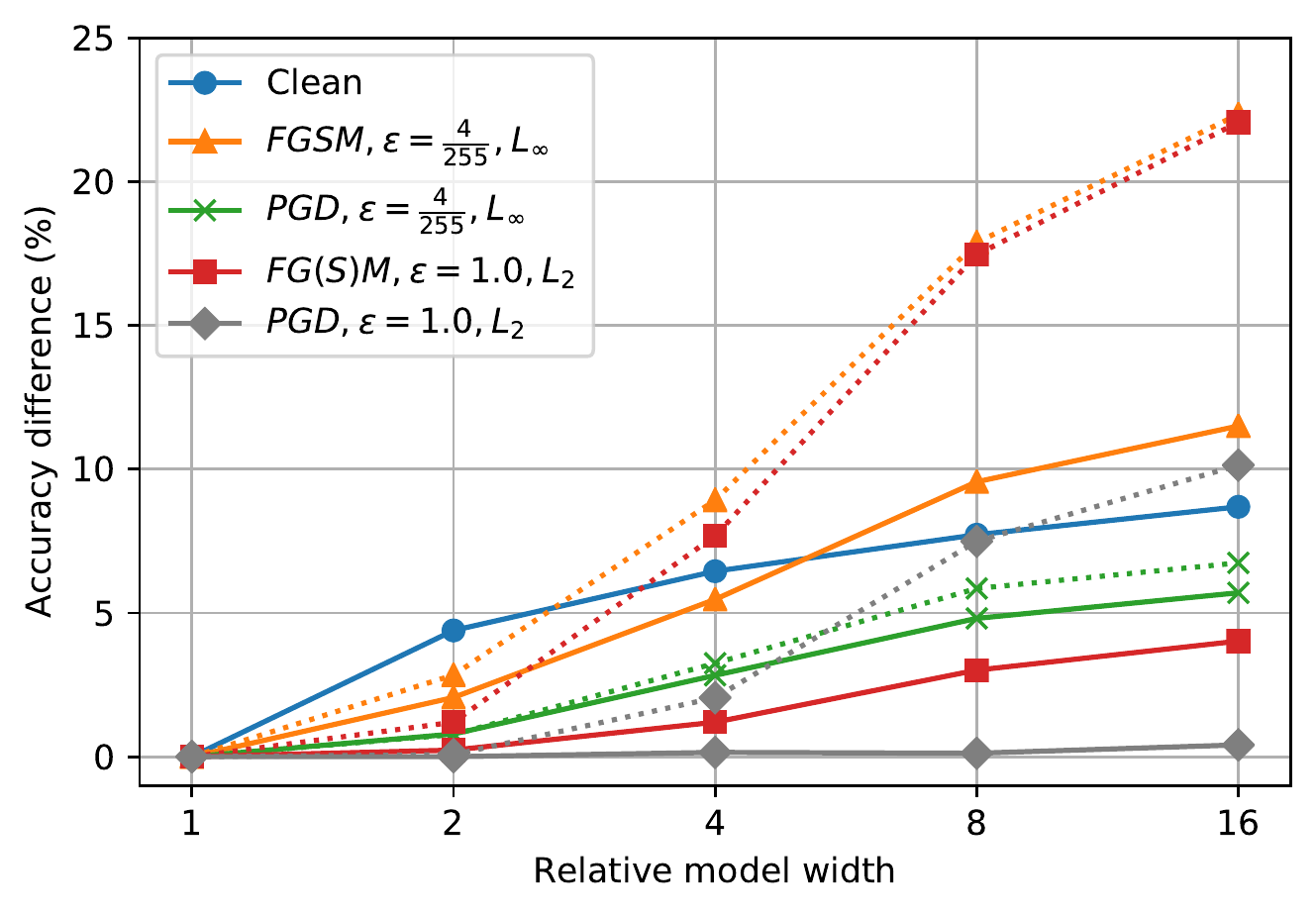}
    \caption{Comparison of clean and adversarial accuracy for Simple models with different relative widths (trained on CIFAR-10). We present difference of accuracy of these models with respect to accuracy of the model with width 1. Dashed and solid line represent accuracy against baseline and compensated attacks, respectively.}
    \label{fig:4-1}
    \vspace{-1em}
\end{figure}

%% file: writing/figure-4-2.tex
\begin{figure}[t]
    \centering
    \includegraphics[width=0.9\linewidth]{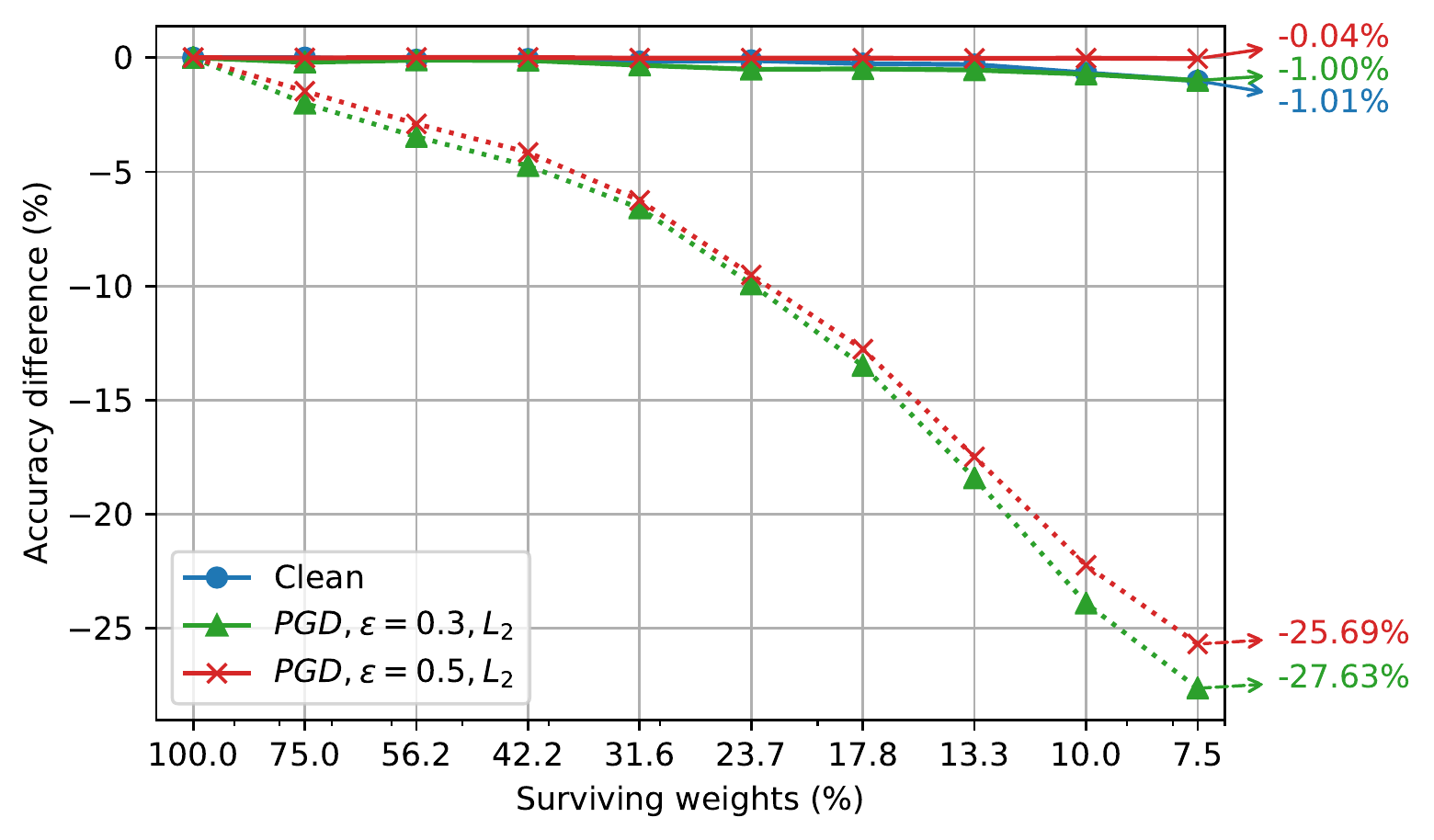}
    \includegraphics[width=0.9\linewidth]{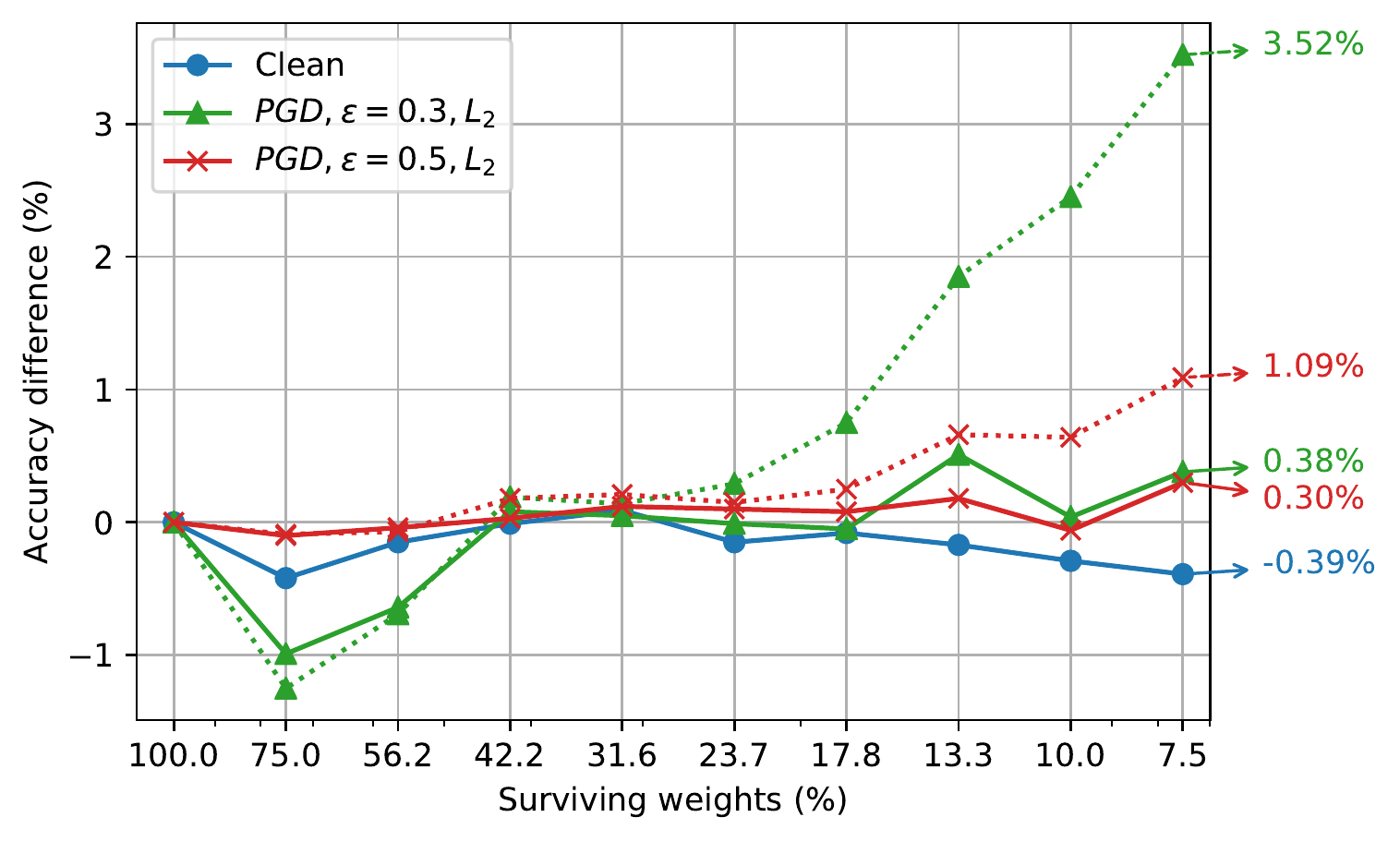}
    \caption{Change of clean and adversarial accuracy through iterative pruning on an over-parameterized WRN 28 (trained on CIFAR-10), trained and finetuned without explicit regularization (above) and with weight decay (below). Dashed and solid lines represent accuracy against baseline attacks and compensated attacks, respectively. The total number of back-propagations is fixed to 9.}
    \label{fig:4-2}
    \vspace{-1em}
\end{figure}

%% file: writing/4-1-1.tex
First, we consider comparing adversarial accuracy among models with the same architecture but with different width (number of output neurons in each layer). Several studies postulate that models with larger width provide better adversarial robustness \cite{madry2017towards, deng2019architecture}, by showing better adversarial accuracy of those models against FGSM or PGD. We examine this claim by measuring adversarial accuracy of Simple models with different relative width trained on CIFAR-10 with fixed weight decay, against both baseline and compensated attacks under the same number of evaluations (i.e., random starts) and back-propagations (i.e., gradient computations for either iterations of PGD or initialization of methods in Sec 3.3) (Figure \ref{fig:4-1}). We find that although models with larger width indeed show better adversarial accuracy, their benefit over smaller models could have been overstated; for example, a PGD attack ($\epsilon=1.0$, $L_2$) gives $10.14\%$ difference in adversarial accuracy between the models with width 1 and 16, but compensating for zero loss and non-differentiability results in only $0.41\%$ difference. Although for other attacks we observe less extreme gaps between baseline and compensated attacks, we generally find that adversarial accuracy of models with larger width tends to be overestimated especially due to the zero loss phenomenon.

%% file: writing/4-1-2.tex
As another approach to benchmark the trade-off between model capacity and adversarial robustness, we consider pruning an over-parameterized model and its effect on adversarial accuracy. We iteratively prune weights of a large WRN 28 (trained on CIFAR-10; details in Appendix C.2) along with finetuning \cite{han2015deep}, both with and without fixed weight decay, and measure adversarial accuracy as in Section 4.1.1 (Figure \ref{fig:4-2}). \emph{Without} weight decay during training and finetuning, adversarial accuracy against baseline attacks drops significantly ($>25\%$) as more weights are pruned away. However, applying compensation methods show that adversarial accuracy actually drops less than $1\%$, similar to that of clean accuracy. The major source of discrepancy is the zero loss phenomenon, in that the original dense model's adversarial accuracy is overestimated due to this phenomenon and the pruned sparse models in fact rely less severely on this phenomenon. 

On the other hand, adversarial accuracy against baseline attacks increases by $3.5\%$ through iterative pruning \emph{with} weight decay, but compensating results in less than $0.4\%$ of increase. We find that weight decay used during finetuning, which adds a large number of epochs (e.g., 10 epochs for finetuning $\times$ 10 pruning iteration $\to$ 100 additional epochs), can act similar to excessive weight decay discussed in Section 3.3. As a result, sparser models show higher adversarial accuracy against baseline PGD, and using initialization methods of Section 3.3 (e.g., PGD + Eigen) gives a more accurate evaluation in this scenario. 

This example illustrates how compensating for the phenomena discussed in Section 3 can prevent misleading conclusions. For example, for WRN 28 models and $L_2$ PGD attacks we tested here, pruning does \emph{not} affect adversarial accuracy significantly. However, without proper compensation, one might conclude that pruning negatively affects adversarial accuracy only observing the case without weight decay, or that pruning improves adversarial accuracy while also reducing the model size after experimenting with weight decay.

%% file: writing/4-2.tex
\subfile{writing/table-4-3.tex}

We compare popular regularization techniques proposed for better generalization or robustness, on whether their adversarial accuracy is affected by the three phenomena we discussed. We train WRN 28 models with different regularization techniques on CIFAR-10, and report adversarial accuracy against baseline and compensated attacks (Table \ref{table:4-2}; details of hyperparameter in Appendix C.3). We observe that input gradient regularization \cite{ross2017improving} and adversarial training \cite{madry2017towards} are least affected by compensation methods, indicating that their robustness does not rely on the phenomena discussed in Section 3. Adversarial accuracy of other regularization methods, such as weight decay and spectral normalization \cite{miyato2018spectral}, partly seems to be overestimated by those phenomena although some of them show better adversarial accuracy even after compensation (e.g., a model trained with spectral normalization shows 21.64\% and 5.25\% decrease in adversarial accuracy for FGSM and PGD, respectively, when compensation methods are applied). Nevertheless, these regularization methods show different behavior for other model architectures and datasets, and those cases are reported in Appendix C.3. 

%% file: writing/table-4-3.tex
\begin{table}[t]
\centering
\caption{Clean and adversarial (FGSM/PGD, $\epsilon=\frac{4}{255}, L_\infty$, total 9 back-propagations for PGD) accuracy (\%) of WRN 28 models trained using different regularization techniques on CIFAR-10. We compare adversarial accuracy against baseline and compensated attacks.}
\vspace{0.5em}
\small
\begin{threeparttable}
\begin{tabular}{@{}cccc@{}}
\toprule
Regularization        & Clean & Attack Baseline & Compensated   \\ \midrule
None                  & 91.65 & 21.90 / 0.02    & 5.94 / 0      \\
Weight decay          & 93.64 & 31.47 / 0.02    & 20.31 / 0.02  \\
Weight decay excess   & 91.52 & 47.83/ 1.85     & 36.46 / 0.63  \\
Spectral norm\tnote{1}         & 87.34 & 31.50 / 7.50    & 9.86 / 2.25   \\
Orthonormal\tnote{2}           & 93.47 & 27.19 / 0.03    & 13.25 / 0.01  \\
Input gradient\tnote{3}              & 89.75 & 24.93 / 12.80   & 23.18 /12.79  \\
Adversarial training\tnote{4}  & 82.67 & 67.62 / 65.58   & 66.46 / 64.82 \\ \bottomrule
\end{tabular}
\begin{tablenotes}\footnotesize
\item[1] Sets the largest eigenvalue of each weight matrix to be 1, following \citeauthor{miyato2018spectral} (\citeyear{miyato2018spectral})
\item[2] Penalizes non-orthonormal weight matrices (\emph{not} full Parseval training of \citeauthor{cisse2017parseval} (\citeyear{cisse2017parseval})), as used in \citeauthor{lin2018defensive} (\citeyear{lin2018defensive})
\item[3] Penalizes the gradients of loss w.r.t. inputs \cite{ross2017improving}
\item[4] Training on adversarial examples of PGD ($\epsilon=\frac{8}{255}, L_\infty$) \cite{madry2017towards}
\end{tablenotes}
\end{threeparttable}
\label{table:4-2}
\end{table}

%% file: writing/5-blackbox.tex
We focus on a black-box attack scenario in which adversarial examples are crafted using a surrogate model that is trained on the same dataset as a target model, but without access to parameters of a target model. We observe that the three phenomena inflating adversarial accuracy discussed in Section 3 can also affect evaluations in black-box setting when a surrogate model suffers from those phenomena. We also show that compensation methods produce transferable adversarial examples, partly accounting for overestimated black-box adversarial accuracy. 

We first examine transferability of examples generated using each compensation method. For the zero loss phenomenon, we find that rescaling logits can provide examples with better transferability, especially when we assume attacks cannot access pre-softmax logits of a target model. Also, examples generated with initialization methods of Section 3.3 only show limited transferability, resulting in less than $<1\%$ of difference for black-box adversarial accuracy when using PGD + Eigen. 

To illustrate transferability, we measure average accuracy against black-box attacks among WRN 28 models trained on CIFAR-10 using different regularization techniques as in Secton 4.2, by using one of them as a surrogate and measuring accuracy on others. On average, compensating for zero loss and non-differentiability gave 11.16\% and 2.90\% difference when the models with no explicit regularization and excessive weight decay are used as surrogate models, respectively, for a PGD attack ($\epsilon=0.5, L_2$, fixed to total 9 back-propagations). When the model with excessive weight decay is a surrogate, using PGD + Eigen accounted for an additional 0.88\% on average. More results on transferability is shown in Appendix D.

%% file: writing/6-verification.tex
\subfile{writing/table-6.tex}

Recently proposed methods for provably robust adversarial training \cite{wong2017provable, sinha2017certifying, raghunathan2018certified} provide the guaranteed lower bound of robustness, although computational efficiency and scalability are challenging issues. Nevertheless, there is usually a gap between the guaranteed lower bound and adversarial accuracy that serves as the natural upper bound of robustness. We investigate whether this gap can be explained by the three phenomena responsible for overestimated adversarial accuracy. 

We experiment on models from \citeauthor{wong2017provable} (\citeyear{wong2017provable}) that are trained to be provably robust, and obtain lower bounds of those models with the verification approach based on MILP \cite{tjeng2018evaluating} that produces tight bounds for deep neural networks with ReLU activation. Details of these models are explained in Appendix E. Then, we compare adversarial accuracy against baseline and compensated PGD (Table \ref{table:6}). We find that the three phenomena at least partially explain the gap between empirical adversarial accuracy and verified lower bounds; compensation methods result in $0.58-1.61\%$ difference in adversarial accuracy, and can bring adversarial accuracy within $0.46\%$ of the lower bound (e.g., CIFAR-B). The major source contributing to this gap is non-differentiability for models trained on MNIST, and zero loss for those trained on CIFAR-10.

%% file: writing/table-6.tex
\begin{table}[t]
\centering
\caption{Comparison of lower bounds of robustness obtained with MILP \cite{tjeng2018evaluating} and empirical adversarial accuracy against both baseline and compensated PGD attacks (5 random starts for both; the total number of back-propagations is 50 for MNIST and 10 for CIFAR-10). Models are trained to be provably robust \cite{wong2017provable} in stated $\epsilon$-ball for $L_\infty$ norm. For each model, attacks use the same $\epsilon$ the model has been trained for as the maximum perturbation size.}
\vspace{0.5em}
\small
\begin{threeparttable}
\begin{tabular}{@{}cccc@{}}
\toprule
\multirow{2}{*}{Model} & \multirow{2}{*}{Lower bound} & \multicolumn{2}{c}{Adversarial accuracy} \\ \cmidrule(l){3-4} 
                       &                              & Baseline    & Compensated       \\ \midrule
MNIST-A, $\epsilon=0.4$           & 52.40\tnote{1}                        & 54.96       & 54.09    \\
MNIST-B, $\epsilon=0.3$           & 75.81\tnote{2}                        & 78.96       & 77.35    \\
CIFAR-A, $\epsilon=\frac{2}{255}$         & 49.80\tnote{2}                        & 51.76       & 51.18    \\
CIFAR-B, $\epsilon=\frac{8}{255}$         & 22.40\tnote{2}                        & 23.49       & 22.86    \\ \bottomrule
\end{tabular}
\begin{tablenotes}\footnotesize
\item[1] Exact robustness obtained with MILP \cite{tjeng2018evaluating}
\item[2] Values directly taken from \citeauthor{tjeng2018evaluating} (\citeyear{tjeng2018evaluating})
\end{tablenotes}
\end{threeparttable}
\label{table:6}
\end{table}

%% file: writing/7-related.tex
\textbf{Attack methods} In this work, we use FGSM, R-FGSM, and PGD for the analysis on their failure cases. There are notable modifications to PGD, such as Basic Iterative Method \cite{kurakin2016adversarial} that omits random initialization of PGD or Momentum Iterative Method \cite{dong2017boosting} that updates each iteration using the momentum term. Our compensation methods are based on these bounded first-order attacks, and provide more accurate gradient computation (Sec 3.1, 3.2) and efficient initialization for PGD (Sec 3.3). Although not examined in this work, Jacobian Saliency Map Attack \cite{Papernot_2016_jsma} provides $L_0$ norm attack method and uses gradients to find each pixel's importance. Alternatively, unbounded attack methods find adversarial examples with minimum perturbation size with optimization-based approaches \cite{szegedy2013intriguing, Carlini_2017, chen2017ead}. For unbounded attack methods, empirical evaluation typically compares the average perturbation size instead of measuring adversarial accuracy. 

\textbf{Gradient masking} \citeauthor{papernot_gradmask} (\citeyear{papernot_gradmask}) observed that black-box attacks can break defensive distillation \cite{Papernot_2016}, which provided robustness against white-box attacks, and called the phenomenon ``gradient masking". Furthermore, \citeauthor{Carlini_2017} (\citeyear{Carlini_2017}) analyzed that temperature softmax used in defensive distillation led to large margin in pre-softmax logits, resulting in zero loss. More generally, \citeauthor{athalye2018obfuscated} (\citeyear{athalye2018obfuscated}) found that certain defenses claiming robustness against white-box attacks relied on obfuscated gradients. Our work is motivated by these previous studies on gradient masking, and contributes by identifying how similar phenomena, such as zero loss (Sec 3.1) and non-differentiability (Sec 3.2), affect evaluation of broad range of deep neural networks. 

\textbf{Verification methods} Another approach to evaluate robustness is verification, which provides guaranteed lower bounds of robustness using optimization techniques. While verification is significantly more difficult than empirically finding adversarial examples using attack methods, recent work \cite{wong2017provable, sinha2017certifying, raghunathan2018certified} could reduce computational complexity of verification by relaxing non-convexities. Moreover, \citeauthor{tjeng2018evaluating} (\citeyear{tjeng2018evaluating}) proposed to use Mixed-Integer Programming to obtain exact robustness for deep neural networks using ReLU as activation. However, retaining computational efficiency while keeping bounds tight and scaling to complex datasets and architectures remain challenging. Our work is orthogonal to verification methods, and can be used to obtain tighter empirical upper bounds with low computational cost (Sec 6). 

\textbf{Benchmarking robustness} There has been increased interest in understanding adversarial robustness for various design choices, such as architectures \cite{Su2018IsRT, deng2019architecture}, activation quantization \cite{lin2018defensive} for hardware efficiency, and regularization techniques for robustness \cite{madry2017towards, cisse2017parseval}. These studies use empirical evaluation against adversarial attacks for numerical experiments.

%% file: writing/8-Conclusion.tex
Overestimated adversarial accuracy has been mainly investigated for defenses that often explicitly capitalized on obfuscated gradients \cite{athalye2018obfuscated} or zero loss \cite{Papernot_2016, Carlini_2017}. In this work, we demonstrate that sources of overestimated adversarial accuracy exist for many conventionally trained deep neural networks, across different architectures and datasets. The three common cases are 1) \emph{zero loss} that induces gradient computation to be inaccurate due to numerical instability, 2) innate \emph{non-differentiability} of ReLU and max pooling that can ``switch" when perturbations are added, resulting in a different set of effective neurons contributing to the final prediction for back-propagation (to compute gradients) and forward-propagation (perturbations are tested for effectiveness), and 3) requiring \emph{more iterations} to successfully find adversarial examples for models trained with certain conditions, such as excessive application of weight decay, thus inflating adversarial accuracy against iterative attacks with small fixed number of iterations. We analyze consequences of these three cases with experiments on different model capacity and regularization techniques, by comparing adversarial accuracy before and after compensating for these three cases. Moreover, we show how these three cases can influence black-box adversarial accuracy, and partially account for the gap between empirical adversarial accuracy and verified lower bounds of robustness. 

Nevertheless, the three cases we identified might not be an exhaustive list responsible for overestimated adversarial accuracy. We still observe a gap between compensated adversarial accuracy and the exact robustness obtained by MILP \cite{tjeng2018evaluating} for MNIST-A in Table \ref{table:6}. This implies that there might exist other phenomena accounting for this gap or that better compensation methods exist for the cases we proposed here. We think future work on investigating other types of gradient masking and compensation methods will benefit empirical studies by further sharpening the metric for adversarial robustness. Additionally, future work on theoretical analysis of our observations, such as models with larger width tend to rely more on zero loss (Sec 4.1.1) or excessive weight decay makes a WRN 28 model less amenable to first-order approximation (Sec 3.3, 4.1.2), can clarify whether they have fundamental implications or are artifacts.

%% file: app-writing/a-1.tex
Simple and Simple-BN architectures briefly described in Section 2.2 are explained in detail in Table \ref{table:a-1}. For WRN 28, we modify the number of output channels and pooling window size to fit with smaller input dimension of CIFAR-10 and SVHN compared to ImageNet. For VGG and WRN used for TinyImageNet, we use the architecture defined as in TorchVision, and only modify final pooling and fully connected layer's dimension to fit downscaled TinyImageNet ($3\times 64\times 64$ with 200 classes). 

\subfile{app-writing/table-a-1.tex}

%% file: app-writing/table-a-1.tex
% \usepackage{multirow}
\begin{table}[h]
\centering
\caption{Description of neural network architectures used in this paper. Convolution layers are specified as (output channel, input channel, kernel height, kernel width, stride, padding). Maxpool layers are in (kernel height, kernel width, stride, padding), and fully connected (FC) layers are in (output channel, input channel). }
\vspace{0.5em}
\small
\begin{tabular}{@{}c|l@{}}
\toprule
Model Type & Description ($w$: width scaling factor)                                                                                                                                                                                                                                                                                                                                                         \\ \midrule
Simple     & \begin{tabular}[c]{@{}l@{}}Conv1 : ($w\times 8$, 3, 3, 3, 1, 1)\\ Conv2 : ($w\times 8$, $w\times 8$, 3, 3, 1, 1)\\ MaxPool: (2, 2, 2, 0)\\ Conv3 : ($w\times 16$, $w\times 8$, 3, 3, 1, 1)\\ Conv4 : ($w\times 16$, $w\times 16$, 3, 3, 1, 1)\\ MaxPool: (2, 2, 2, 0)\\ FC1 : ($w\times 128$, $w\times 16\times 8\times 8$)\\ FC2 : (10, $w\times 128$)\end{tabular} \\ \midrule
Simple-BN  & \begin{tabular}[c]{@{}l@{}}Convolution and FC layers are same as in Simple, \\ but Batch Normalization layer follows each \\ Convolution layer.\end{tabular}                                                                                                                                                                                                    \\ \bottomrule
\end{tabular}
\label{table:a-1}
\end{table}

%% file: app-writing/a-2.tex
As a default setting, we train for 100 epochs using Stochastic Gradient Descent (SGD) with momentum of 0.9 for CIFAR-10 and SVHN. For Simple-BN and WRN 28, we use starting learning rate of 0.1 and decay it by factor of 10 for every 40 epochs. For Simple, we start with learning rate of 0.01. Models for TinyImageNet are trained with Adam ($\beta_1=0.9, \beta_2=0.99$), with starting learning rate of 0.001. Learning rate decay is applied in the same manner. Default batch size is 128, unless GPU memory is insufficient. Different training conditions deviating from the default setting, including specific regularizations, are described when they are introduced in Appendix C. 

%% file: app-writing/a-3.tex
Fast Gradient Sign Method (FGSM) for $L_\infty$ norm with the perturbation size $\epsilon$ can be expressed as:
\begin{equation}
    x_{adv} = x + \epsilon\cdot \texttt{sign}(g) = x + \epsilon\cdot\texttt{sign}(\frac{\partial l(f(x), t)}{\partial x})
\end{equation}
Then, Random-FGSM (R-FGSM) modifies FGSM by adding a random perturbation before computing gradients:
\begin{gather}
    x' = x + \frac{\epsilon}{2}\cdot \texttt{sign}(r) \label{eq2}\\ 
    x_{adv} = x' + \frac{\epsilon}{2}\cdot\texttt{sign}(\frac{\partial l(f(x'), t)}{\partial x'})
\end{gather}
where $r$ is randomly sampled from normal distribution $\mathcal{N}(0, I)$. Projected Gradient Descent (PGD) updates $x_{adv}$ iteratively, and the equation for step $s+1$ can be expressed as:
\begin{gather}
    x_{adv}^{(0)} = x + \epsilon\cdot\texttt{sign}(r) \label{eq4}\\
    x_{adv}^{(s+1)} = x_{adv}^{(s)} + \texttt{clip}(\alpha\cdot\texttt{sign}(\frac{\partial l(f(x_{adv}^{(s)}, t)}{\partial x_{adv}^{(s)}}), \ \epsilon)
\end{gather}
where $r$ is random initialization vector as in the equation \eqref{eq2}, and $\alpha$ is a step size. \texttt{Clip} operation ensures the perturbation to be within the $\epsilon$-ball around the sample $x$. For $L_2$ norm, \texttt{sign} operation is replaced with dividing by the size $\|g\|_2$ to obtain a unit vector. That is, for FGSM:
\begin{equation}
    x_{adv} = x + \epsilon\cdot\frac{g}{\|g\|_2}
\end{equation}
Other attacks can be similarly modified. Since \texttt{sign} operation no longer exists, we can drop `S' from names of attacks (e.g., FG(S)M and R-FG(S)M). 

%% file: app-writing/b-1-1.tex
First, we train a Simple model ($w=4$) without explicit regularization on CIFAR-10, and analyze the average loss and size of gradients (in $L_2$ norm) (Table \ref{table:b-1-1}). We characterize those statistics for samples on which a FGSM attack ($\epsilon=\frac{8}{255}$) succeeds and fails. Observe that samples on which the attack fails have smaller loss and size of gradients induced by large logit margin. 
\subfile{app-writing/table-b-1-1.tex}

Furthermore, we observe that a black-box attack on the above-mentioned model is stronger than a white-box attack: when the model with the same architecture, width, and initialization but trained independently using weight decay regularization (with hyperparameter $5\times10^{-4}$) is used as a surrogate, a black-box FGSM attack ($\epsilon=\frac{8}{255}$) on the above-mentioned model results in $13.22\%$ accuracy (in other words, $86.78\%$ attack success). However, a white box FGSM attack with the same $\epsilon$ results in $14.04\%$ accuracy, higher than that against the black-box attack. This observation indicates that a black-box attack is stronger than a white-box attack for this model, signaling a possible gradient masking.

Additionally, we visualize how loss changes as a perturbation is added (Fig \ref{fig:b-1-1}) on a randomly chosen clean sample $x$ on which the white-box attack fails but the black-box attack succeeds. This sample has zero cross-entropy loss. We can observe that moving along the direction of $g$ (while $\epsilon_2=0$) does not increase loss even when the perturbation size is increased to $\epsilon_1=\frac{16}{255}$. However, notice that the black-box attack's gradients easily increase loss (Fig \ref{fig:b-1-1} (a)). We also visualize loss when $\epsilon_2$ moves along the direction of gradients produced by changing the target label for loss to be the second most likely class (Fig \ref{fig:b-1-1} (b)). This illustrates how gradients computed from zero loss do not give meaningful perturbation direction, and why failure of attacks in such case does not indicate robustness (i.e., existence of adversarial examples in other perturbation directions, such as gradients of the black-box attack). 
\subfile{app-writing/fig-b-1-1.tex}

%% file: app-writing/table-b-1-1.tex
\begin{table}[h]
\centering
\caption{Characterizing samples on which the white-box attack using FGSM ($\epsilon=\frac{8}{255}$) succeeds and fails for the value of loss, size of gradients, and logit statistics including margin and variance, for a Simple model with width scale factor of $4$ trained without explicit regularization on CIFAR-10.}
\vspace{0.5em}
\small
\begin{tabular}{@{}ccc@{}}
\toprule
               & Attack succeed & Attack fail \\ \midrule
Loss           & 0.0643         & 0.0011      \\
Gradient       & 2.0686         & 0.0339      \\
Logit margin   & 7.20           & 18.73       \\
Logit variance & 73.44          & 142.19      \\ \bottomrule
\end{tabular}
\label{table:b-1-1}
\end{table}

%% file: app-writing/fig-b-1-1.tex
\begin{figure}[h]
    \centering
    \subfigure[]{\includegraphics[width=0.4\linewidth]{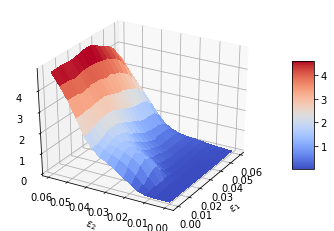}} \quad
    \subfigure[]{\includegraphics[width=0.4\linewidth]{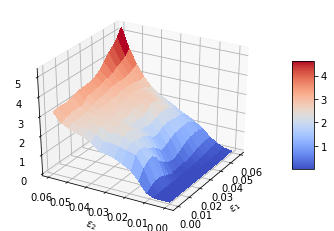}}
    \caption{Visualization of the value of loss (z-axis) evaluated for points $x^* = x + \epsilon_1\cdot\texttt{sign}(g) + \epsilon_2\cdot\texttt{sign}(g^\prime)$, where $g$ is gradients of loss with respect to the input sample computed using the target model itself, and $g^\prime$ is (a) gradients computed from a surrogate model used for a black-box attack, and (b) gradients computed using the second most likely class as the label for loss instead of the ground truth.}
    \label{fig:b-1-1}
\end{figure}

%% file: app-writing/b-1-2.tex
\subfile{app-writing/table-b-1-2.tex}

We compare the effectiveness of compensation methods discussed in Section 3.1, by comparing accuracy against compensated attack methods on a Simple model trained without explicit regularization on CIFAR-10 (Table \ref{table:b-1-2}). Generally, changing the target label to be the second most likely class (corresponding to the second largest logit) gives the best compensation for this model. 

%% file: app-writing/table-b-1-2.tex
\begin{table}[h]
\centering
\caption{Adversarial accuracy (in \%) for different attack types compensated for the zero loss phenomenon. We sweep a constant $T$ for rescaling logits, and target labels $t$ for changing loss computation; `random' sets $t$ to be randomly sampled among all possible classes, `least' and `second' indicate the least-likely and the second most-likely classes, respectively. We also state the gap between baseline and compensated adversarial accuracy.}
\vspace{0.5em}
\small
\begin{tabular}{@{}c|c|c|ccc|ccc|c@{}}
\toprule
\multicolumn{2}{c|}{}                                         &                           & \multicolumn{3}{c|}{Rescaling} & \multicolumn{3}{c|}{Targeted}   & {\color[HTML]{FE0000} }                      \\
\multicolumn{2}{c|}{\multirow{-2}{*}{Evaluation type}}        & \multirow{-2}{*}{Baseline} & $T=10$   & $T=50$  & $T=100$  & Random & Least & Second         & \multirow{-2}{*}{{\color[HTML]{FE0000} Gap}} \\ \midrule
\multicolumn{2}{c|}{Clean}                                    & 84.75                     & -        & -       & -        & -      & -     & -              & {\color[HTML]{FE0000} -}                     \\ \midrule
                                                     & FGSM   & 19.50                     & 13.77    & 15.24   & 15.38    & 14.59  & 17.62 & \textbf{10.79} & {\color[HTML]{FE0000} 8.71}                  \\
                                                     & R-FGSM & 29.81                     & 30.35    & 30.44   & 30.46    & 30.02  & 30.73 & \textbf{29.11} & {\color[HTML]{FE0000} 0.70}                  \\
\multirow{-3}{*}{$L_\infty, \epsilon=\frac{4}{255}$} & PGD    & 2.55                      & 2.06     & 2.26    & 2.26     & 2.03   & 2.67  & \textbf{1.68}  & {\color[HTML]{FE0000} 0.87}                  \\ \midrule
                                                     & FGSM   & 14.04                     & 3.51     & 4.40    & 4.51     & 4.20   & 6.52  & \textbf{3.08}  & {\color[HTML]{FE0000} 10.96}                 \\
                                                     & R-FGSM & 15.03                     & 9.58     & 11.29   & 11.50    & 11.12  & 13.78 & \textbf{7.23}  & {\color[HTML]{FE0000} 7.80}                  \\
\multirow{-3}{*}{$L_\infty, \epsilon=\frac{8}{255}$} & PGD    & 0.24                      & 0        & 0.01    & 0.01     & 0.01   & 0.08  & \textbf{0   }  & {\color[HTML]{FE0000} 0.24}                  \\ \midrule
                                                     & FGM    & 23.18                     & 20.45    & 21.46   & 21.54    & 20.77  & 22.67 & \textbf{18.10} & {\color[HTML]{FE0000} 5.08}                  \\
                                                     & R-FGM  & 39.64                     & 40.09    & 40.23   & 40.26    & 39.81  & 40.33 & \textbf{39.08} & {\color[HTML]{FE0000} 0.56}                  \\
\multirow{-3}{*}{$L_2, \epsilon=0.5$}                & PGD    & 6.67                      & 5.06     & 5.68    & 5.75     & 5.25   & 6.85  & \textbf{3.83}  & {\color[HTML]{FE0000} 2.84}                  \\ \midrule
                                                     & FGM    & 17.75                     & 7.75     & 9.37    & 9.56     & 9.03   & 12.32 & \textbf{6.14}  & {\color[HTML]{FE0000} 11.61}                 \\
                                                     & R-FGM  & 19.84                     & 17.88    & 18.96   & 19.04    & 18.36  & 20.32 & \textbf{15.24} & {\color[HTML]{FE0000} 4.60}                  \\
\multirow{-3}{*}{$L_2, \epsilon=1.0$}                & PGD    & 3.82                      & 0.11     & 0.28    & 0.28     & 0.25   & 1.39  & \textbf{0.03}  & {\color[HTML]{FE0000} 3.79}                  \\ \bottomrule
\end{tabular}
\label{table:b-1-2}
\end{table}

%% file: app-writing/b-1-3.tex
\subfile{app-writing/table-b-1-3.tex}
In addition to comparing models trained with and without weight decay for how they are affected by the zero loss phenomenon, as in Section 3.1, we additionally compare models with larger width or batch normalization (Table \ref{table:b-1-3}). For the Simple architecture and CIFAR-10 dataset we investigate here, models with larger width or batch normalization are affected more severely than a model with smaller width and no batch normalization. 

%% file: app-writing/table-b-1-3.tex
\begin{table}[h]
\centering
\caption{The gap in adversarial accuracy (in \%) when compensating for the zero loss phenomenon by changing target labels to be the second most likely classes. `No Reg' represents a Simple model with width scale factor of $4$ trained without explicit regularization. `Weight Decay' represents the same model as `No Reg', but trained with weight decay of $5\times 10^{-4}$. `Width x4' represents a model with 4 times more neurons per layer compared to `No Reg', thus is a Simple model with width scale factor of $16$. `Batch Norm' indicates a Simple-BN model, which is same as `No Reg' except for batch normalization following each convolutional layer. }
\vspace{0.5em}
\small
\begin{tabular}{@{}c|c|cccc@{}}
\toprule
\multicolumn{2}{c|}{Evaluation type}                    & No Reg & Weight Decay & Width x4 & Batch Norm \\ \midrule
\multirow{3}{*}{$L_\infty, \epsilon=\frac{4}{255}$} & FGSM   & 8.71   & 4.48         & 19.48    & 18.73      \\
                                                    & R-FGSM & 0.70   & 0.59         & 5.78     & 9.36         \\
                                                    & PGD    & 0.87   & 0.19         & 4.51     & 6.16       \\ \midrule
\multirow{3}{*}{$L_\infty, \epsilon=\frac{8}{255}$} & FGSM   & 10.96  & 8.12         & 18.43    & 19.99      \\
                                                    & R-FGSM & 7.80   & 3.89         & 17.75    & 15.68      \\
                                                    & PGD    & 0.24   & 0.15         & 1.20     & 2.05       \\ \midrule
\multirow{3}{*}{$L_2, \epsilon=0.5$}                & FGM    & 5.08   & 2.41         & 16.54    & 16.78      \\
                                                    & R-FGM  & 0.56   & 0.63         & 2.72     & 5.64       \\
                                                    & PGD    & 2.84   & 0.57         & 15.29    & 12.68      \\ \midrule
\multirow{3}{*}{$L_2, \epsilon=1.0$}                & FGM    & 11.61  & 7.91         & 27.34    & 21.02      \\
                                                    & R-FGM  & 4.60   & 1.73         & 16.24    & 15.51      \\
                                                    & PGD    & 3.79   & 2.14         & 19.62    & 9.45      \\ \bottomrule
\end{tabular}
\label{table:b-1-3}
\end{table}

%% file: app-writing/b-2.tex
Here we present different choices of substitute functions and their effectiveness. First, as a substitute for ReLU, we consider softplus ($\alpha=2$, threshold set to 2), CELU ($\alpha=2$), and ELU ($\alpha=1$). While softplus and CELU are continuously differentiable at zero regardless of the hyperparameter $\alpha$ that controls the slope, $\alpha$ for ELU has to be fixed to 1 to ensure differentiability at zero. Second, if a model uses max pooling, we substitute max pooling with $L_p$ norm pooling. As $p \to \infty$, $L_p$ norm pooling gets closer to max pooling. Thus, sufficiently large but finite $p$ can provide differentiable approximation of max pooling. Tables \ref{table:b-2-1} and \ref{table:b-2-2} show the effectiveness of each choice for a Simple model and a WRN 28 model. Tables \ref{table:b-2-3} and \ref{table:b-2-4} compare each choice when combined with zero loss compensation. We observe that using softplus and $p=5, 10$ generally gives the most decrease in adversarial accuracy, when compensating for both this and the zero loss phenomena. However, when only compensating for this phenomenon, other substitute functions often give better compensation. 

\subfile{app-writing/table-b-2-1.tex}
\subfile{app-writing/table-b-2-2.tex}
\subfile{app-writing/table-b-2-3.tex}
\subfile{app-writing/table-b-2-4.tex}

%% file: app-writing/table-b-2-1.tex
\begin{table}[h]
\centering
\caption{Adversarial accuracy (in \%) of a Simple model, same as in Table \ref{table:b-1-2}, for different attack types compensated for innate non-differentiability using BPDA. We investigate three differentiable functions for substituting ReLU: Softplus ($\alpha=2$, thresholded at 2), CELU ($\alpha=2$), and ELU ($\alpha=1$) while fixing $L_p$-norm pool's $p$ to be 5. Then, we sweep for $p$ by fixing ReLU substitute function to be the one achieved the best performance. }
\vspace{0.5em}
\small
\begin{tabular}{@{}c|c|c|ccc|ccc|c@{}}
\toprule
\multicolumn{2}{c|}{}                                         &                           & \multicolumn{3}{c|}{ReLU Substitute} & \multicolumn{3}{c|}{$L_p$-norm Pool} & {\color[HTML]{FE0000} }                      \\
\multicolumn{2}{c|}{\multirow{-2}{*}{Evaluation Type}}                       & \multirow{-2}{*}{Baseline} & Softplus      & CELU       & ELU        & $p=2$   & $p=5$   & $p=10$  & \multirow{-2}{*}{{\color[HTML]{FE0000} Gap}} \\ \midrule
\multicolumn{2}{c|}{Clean}                                    & 84.75                     & -             & -          & -          & -       & -       & -       & {\color[HTML]{FE0000} -}                     \\ \midrule
                                                     & FGSM   & 19.50                     & \textbf{18.41}         & 18.93      & 18.68      & 18.58   & \textbf{18.41}   & 18.44   & {\color[HTML]{FE0000} 1.09}                  \\
                                                     & R-FGSM & 29.81                     & \textbf{29.17}         & 30.51      & 29.92      & 30.26   & 29.17   & \textbf{29.15}   & {\color[HTML]{FE0000} 0.66}                  \\
\multirow{-3}{*}{$L_\infty, \epsilon=\frac{4}{255}$} & PGD    & 2.55                      & 2.54          & 2.54       & \textbf{2.50}       & 2.57    & 2.50    & \textbf{2.47}    & {\color[HTML]{FE0000} 0.08}                  \\ \midrule
                                                     & FGSM   & 14.04                     & 12.33         & \textbf{12.06}      & 12.15      & \textbf{11.85}   & 12.06   & 12.11   & {\color[HTML]{FE0000} 2.19}                  \\
                                                     & R-FGSM & 15.03                     & 14.88         & 14.95      & \textbf{14.72}      & 15.03   & \textbf{14.72}   & 14.78   & {\color[HTML]{FE0000} 0.31}                  \\
\multirow{-3}{*}{$L_\infty, \epsilon=\frac{8}{255}$} & PGD    & 0.24                      & 0.21          & \textbf{0.15}       & 0.16       & 0.22    & 0.15    & \textbf{0.13}    & {\color[HTML]{FE0000} 0.11}                  \\ \midrule
                                                     & FGM    & 23.18                     & \textbf{21.07}         & 22.43      & 22.22      & 21.66   & 21.07   & \textbf{20.59}   & {\color[HTML]{FE0000} 2.59}                  \\
                                                     & R-FGM  & 39.64                     & \textbf{38.85}         & 40.05      & 39.72      & 39.77   & 38.85   & \textbf{35.67}   & {\color[HTML]{FE0000} 3.97}                  \\
\multirow{-3}{*}{$L_2, \epsilon=0.5$}                & PGD    & 6.67                      & 6.68          & 6.57       & \textbf{6.55}       & 6.56    & \textbf{6.55}    & 6.64    & {\color[HTML]{FE0000} 0.12}                  \\ \midrule
                                                     & FGM    & 17.75                     & 16.07         & \textbf{16.04}      & 16.25      & 15.87   & 16.04   & \textbf{15.78}   & {\color[HTML]{FE0000} 1.97}                  \\
                                                     & R-FGM  & 19.84                     & \textbf{18.69}         & 19.66      & 19.39      & 19.06   & 18.69   & \textbf{17.98}   & {\color[HTML]{FE0000} 1.86}                  \\
\multirow{-3}{*}{$L_2, \epsilon=1.0$}                & PGD    & 3.82                      & 3.90          & 3.64       & \textbf{3.63}       & \textbf{3.58}     & 3.63    & 3.67   & {\color[HTML]{FE0000} 0.24}                  \\ \bottomrule
\end{tabular}
\label{table:b-2-1}
\end{table}

%% file: app-writing/table-b-2-2.tex
\begin{table}[h]
\centering
\caption{Adversarial accuracy (in \%) of a WRN 28 (width scale factor: $2$), trained on CIFAR-10 without explicit regularization, compensated for innate non-differentiability using BPDA. Details are same as in Table \ref{table:b-2-1}, except for that this model does not use max pooling. }
\vspace{0.5em}
\small
\begin{tabular}{@{}c|c|c|ccc|c@{}}
\toprule
\multicolumn{2}{c|}{}                                         &                           & \multicolumn{3}{c|}{ReLU Substitute} & {\color[HTML]{FE0000} }                      \\
\multicolumn{2}{c|}{\multirow{-2}{*}{Evaluation Type}}        & \multirow{-2}{*}{Baseline} & Softplus     & CELU      & ELU      & \multirow{-2}{*}{{\color[HTML]{FE0000} Gap}} \\ \midrule
\multicolumn{2}{c|}{Clean}                                    & 91.65                     & -            & -         & -        & {\color[HTML]{FE0000} -}                     \\ \midrule
                                                     & FGSM   & 21.90                     & \textbf{15.87}        & 21.36     & 20.06    & {\color[HTML]{FE0000} 6.03}                  \\
                                                     & R-FGSM & 20.79                     & \textbf{19.76}        & 23.71     & 23.30    & {\color[HTML]{FE0000} 1.03}                  \\
\multirow{-3}{*}{$L_\infty, \epsilon=\frac{4}{255}$} & PGD    & 0.04                      & 0.04                  & 0.04      & 0.04     & {\color[HTML]{FE0000} 0.00}                  \\ \midrule
                                                     & FGSM   & 17.57                     & \textbf{10.93}        & 15.38     & 13.83    & {\color[HTML]{FE0000} 6.64}                  \\
                                                     & R-FGSM & 5.41                      & \textbf{4.43}         & 7.28      & 6.50     & {\color[HTML]{FE0000} 0.98}                  \\
\multirow{-3}{*}{$L_\infty, \epsilon=\frac{8}{255}$} & PGD    & 0                         & -            & -         & -        & {\color[HTML]{FE0000} -}                     \\ \midrule
                                                     & FGM    & 47.30                     & 44.73        & 45.46     & \textbf{42.76}    & {\color[HTML]{FE0000} 4.54}                  \\
                                                     & R-FGM  & 45.13                     & \textbf{44.11}        & 46.87     & 45.29    & {\color[HTML]{FE0000} 1.02}                  \\
\multirow{-3}{*}{$L_2, \epsilon=0.5$}                & PGD    & 22.58                     & 21.92        & 18.15     & \textbf{15.78}    & {\color[HTML]{FE0000} 6.80}                  \\ \midrule
                                                     & FGM    & 45.33                     & 42.57        & 40.75     & \textbf{38.26}    & {\color[HTML]{FE0000} 7.07}                  \\
                                                     & R-FGM  & 36.18                     & 35.89        & 37.05     & \textbf{33.78}    & {\color[HTML]{FE0000} 2.40}                  \\
\multirow{-3}{*}{$L_2, \epsilon=1.0$}                & PGD    & 16.29                     & 15.81        & \textbf{11.09}     & 11.95    & {\color[HTML]{FE0000} 5.20}                  \\ \bottomrule
\end{tabular}
\label{table:b-2-2}
\end{table}

%% file: app-writing/table-b-2-3.tex
\begin{table}[h]
\centering
\caption{Adversarial accuracy (in \%) of a Simple model, same as in Table \ref{table:b-2-1}, when compensated for both zero loss and innate non-differentiability. The zero-loss phenomenon is compensated by changing the targe labels to be the second most likely classes.}
\vspace{0.5em}
\small
\begin{tabular}{@{}c|c|c|ccc|ccc|c@{}}
\toprule
\multicolumn{2}{c|}{}                                         &                                  & \multicolumn{3}{c|}{ReLU Substitute} & \multicolumn{3}{c|}{$L_p$-norm Pool}             & {\color[HTML]{FE0000} }                      \\
\multicolumn{2}{c|}{\multirow{-2}{*}{Evaluation Type}}        & \multirow{-2}{*}{Zero Loss Only} & Softplus         & CELU    & ELU     & $p=2$         & $p=5$          & $p=10$         & \multirow{-2}{*}{{\color[HTML]{FE0000} Gap}} \\ \midrule
\multicolumn{2}{c|}{Clean}                                    & 84.75                            & -                & -       & -       & -             & -              & -              & {\color[HTML]{FE0000} -}                     \\ \midrule
                                                     & FGSM   & 10.79                            & \textbf{8.74}    & 10.05   & 9.45    & 9.40          & \textbf{8.74}  & 8.86           & {\color[HTML]{FE0000} 2.23}                  \\
                                                     & R-FGSM & 29.11                            & \textbf{28.31}   & 29.78   & 29.11   & 29.45         & \textbf{28.31} & 28.39          & {\color[HTML]{FE0000} 0.80}                  \\
\multirow{-3}{*}{$L_\infty, \epsilon=\frac{4}{255}$} & PGD    & \textbf{1.68}                    & 1.70             & 1.73    & 1.70    & 1.70          & 1.70           & 1.70           & {\color[HTML]{FE0000} -0.02}                  \\ \midrule
                                                     & FGSM   & 3.08                             & \textbf{1.25}    & 1.62    & 1.56    & \textbf{1.13} & 1.25           & 1.43           & {\color[HTML]{FE0000} 1.95}                  \\
                                                     & R-FGSM & 7.23                             & \textbf{6.25}    & 7.17    & 6.47    & 6.64          & \textbf{6.25}  & 6.28           & {\color[HTML]{FE0000} 0.98}                  \\
\multirow{-3}{*}{$L_\infty, \epsilon=\frac{8}{255}$} & PGD    & 0                               & -                 & -       & -       & -             & -              & -              & {\color[HTML]{FE0000} -  }                  \\ \midrule
                                                     & FGM    & 18.10                            & \textbf{14.94}   & 17.01   & 16.56   & 15.94         & 14.94          & \textbf{13.74} & {\color[HTML]{FE0000} 4.36}                  \\
                                                     & R-FGM  & 39.08                            & \textbf{38.29}   & 39.47   & 39.10   & 39.26         & 38.29          & \textbf{34.33} & {\color[HTML]{FE0000} 4.75}                  \\
\multirow{-3}{*}{$L_2, \epsilon=0.5$}                & PGD    & \textbf{3.83}                    & 3.90             & 3.90    & 3.88    & 3.93          & 3.88           & 3.91           & {\color[HTML]{FE0000} -0.05}                 \\ \midrule
                                                     & FGM    & 6.14                             & \textbf{3.11}    & 4.47    & 4.41    & 3.15          & 3.11           & \textbf{3.08}  & {\color[HTML]{FE0000} 3.06}                  \\
                                                     & R-FGM  & 15.24                            & \textbf{13.16}   & 14.95   & 14.48   & 13.90         & 13.16          & \textbf{12.16} & {\color[HTML]{FE0000} 3.08}                  \\
\multirow{-3}{*}{$L_2, \epsilon=1.0$}                & PGD    & \textbf{0.03}                    & 0.03             & 0.03    & 0.03    & 0.03          & 0.03           & 0.04           & {\color[HTML]{FE0000} 0.00}                  \\ \bottomrule
\end{tabular}
\label{table:b-2-3}
\end{table}

%% file: app-writing/table-b-2-4.tex
\begin{table}[h]
\centering
\caption{Adversarial accuracy (in \%) of a WRN 28, same as in Table \ref{table:b-2-2}, compensated for both zero loss and innate nondifferentiability.}
\vspace{0.5em}
\small
\begin{tabular}{@{}c|c|c|ccc|c@{}}
\toprule
\multicolumn{2}{c|}{}                                         &                                  & \multicolumn{3}{c|}{ReLU Substitute} & {\color[HTML]{FE0000} }                      \\
\multicolumn{2}{c|}{\multirow{-2}{*}{Evaluation Type}}        & \multirow{-2}{*}{Zero Loss Only} & Softplus         & CELU    & ELU    & \multirow{-2}{*}{{\color[HTML]{FE0000} Gap}} \\ \midrule
\multicolumn{2}{c|}{Clean}                                    & 91.65                            & -                & -       & -      & {\color[HTML]{FE0000} -}                     \\ \midrule
                                                     & FGSM   & 11.34                            & \textbf{5.94}    & 11.08   & 10.58  & {\color[HTML]{FE0000} 5.40}                  \\
                                                     & R-FGSM & 13.46                            & \textbf{11.25}   & 15.21   & 15.01  & {\color[HTML]{FE0000} 2.21}                  \\
\multirow{-3}{*}{$L_\infty, \epsilon=\frac{4}{255}$} & PGD    & 0                                & -                & -       & -      & {\color[HTML]{FE0000} -   }                  \\ \midrule
                                                     & FGSM   & 7.89                             & \textbf{3.36}    & 6.87    & 6.01   & {\color[HTML]{FE0000} 4.53}                  \\
                                                     & R-FGSM & 2.49                             & \textbf{1.3}     & 3.55    & 3.27   & {\color[HTML]{FE0000} 1.19}                  \\
\multirow{-3}{*}{$L_\infty, \epsilon=\frac{8}{255}$} & PGD    & 0                                & -                & -       & -      & {\color[HTML]{FE0000} -}                     \\ \midrule
                                                     & FGM    & 21.32                            & \textbf{11.72}   & 21.1    & 20.49  & {\color[HTML]{FE0000} 9.60}                  \\
                                                     & R-FGM  & 27.74                            & \textbf{21.93}   & 30.15   & 29.4   & {\color[HTML]{FE0000} 5.81}                  \\
\multirow{-3}{*}{$L_2, \epsilon=0.5$}                & PGD    & 0                                & -                & -       & -      & {\color[HTML]{FE0000} -   }                  \\ \midrule
                                                     & FGM    & 15.55                            & \textbf{7.3}     & 14.59   & 13.62  & {\color[HTML]{FE0000} 8.25}                  \\
                                                     & R-FGM  & 12.51                            & \textbf{7.73}    & 14.91   & 14.39  & {\color[HTML]{FE0000} 4.78}                  \\
\multirow{-3}{*}{$L_2, \epsilon=1.0$}                & PGD    & 0                                & -                & -       & -      & {\color[HTML]{FE0000} -}                     \\ \bottomrule
\end{tabular}
\label{table:b-2-4}
\end{table}

%% file: app-writing/b-3.tex
\textbf{Eigenvector approximation (PGD + Eigen)} \citeauthor{Miyato_2019} (\citeyear{Miyato_2019}) proposed a method to approximate the principle eigenvector $u$ of the Hessian $H$ for a semi-supervised learning problem. Although their objective function for which they computed $u$ and $H$ is different from ours, we adopt their general principle to use power iteration and finite difference method to approximate $u$ of our $H=\nabla\nabla_x l(f(x), t)$. Power iteration starts with a randomly sampled vector $d$, with an assumption that $d$ has non-zero projection on the principle eigenvector $u$. Then, $d$ is updated as $d \leftarrow \frac{H\cdot d}{\|H\cdot d\|_2}$ until it converges. Finite difference method is used to approximate $H\cdot d$ when exact $H$ is hard to obtain; since only the product $H\cdot d$ is necessary, finite difference method computes the difference between first-order derivatives measured for $x$ and $x+\delta\cdot d$ ($\delta > 0$) as an approximation for $H\cdot d$.

Furthermore, we use $u$ produced by this approximation method as a direction for \emph{initialization} instead of the perturbation $r$ itself. The motivation behind this approach is to initialize PGD's first-order iterations to a point with high curvature, so that gradients and loss can increase rapidly (Sec 3.3). Since $u$ is a unit vector in $L_2$ norm, we use $\epsilon\cdot u$ as initialization for PGD in $L_2$ norm, and $\texttt{clip}(\sqrt{\frac{d_x}{\pi}}\cdot\epsilon\cdot u, \texttt{ min}=-\epsilon, \texttt{ max}=\epsilon)$ for $L_\infty$ norm where $d_x$ is the dimension of input samples. Note that this method requires back-propagation twice when approximating $H\cdot d$ using finite difference method. 

\textbf{Quasi-Newton method (PGD + BFGS)} BFGS (Broyden–Fletcher–Goldfarb–Shanno algorithm) is a second-order optimization method that approximates the Quasi-Newton direction for update $H^{-1}\cdot g$. Since we are only interested in the direction for initialization, rather than fully iterate with second-order method, we only consider approximating the inverse of the Hessian $H^{-1}$ and omit line search to obtain step size. We use a single update for $H^{-1}$:
\begin{gather}
    \Delta_g = \frac{\partial l(f(x^\prime), t)}{\partial x^\prime}\Bigr|_{x^\prime=x+d} - \frac{\partial l(f(x^\prime), t)}{\partial x^\prime}\Bigr|_{x^\prime=x} \\
    H^{-1} = (I-\frac{d\Delta_g^T}{\Delta_g^T d})(I-\frac{\Delta_g d^T}{\Delta_g^T d}) + \frac{dd^T}{\Delta_g^T d} \label{eq:b-2}
\end{gather}
where $d$ is randomly sampled from $\mathcal{N}(0, I)$ and scaled to have a small size $\delta$. Note that the update in Eq \eqref{eq:b-2} is same as typical BFGS inverse update for the first iteration where $H^{-1}_{0}$ is initialized as the identity matrix. Then, we use $H^{-1}\cdot g=H^{-1}\cdot\frac{\partial l(f(x), t)}{\partial x}$ as the direction for initialization.

Similar to PGD + Eigen, this method requires two additional back-propagations for computing $\Delta_g$. However, this method consumes more memory as it directly computes $H^{-1}$ instead of the matrix-vector product $H\cdot d$. Since $H^{-1}$ is a $d_x \times d_x$ matrix, batch size might have to be reduced to accommodate significant additional memory of $H^{-1}$. 

%% file: app-writing/c.tex
Accuracy against compensated attacks is measured in a cascading manner, in which samples that survived the previous stage (i.e., an attack fails on that sample to find adversarial perturbation) are subjected to the next compensation method. For a single-step attack, such as FGSM or R-FGSM, we cascade compensation methods in following steps:
\begin{enumerate}
    \item Apply an attack without any compensation methods. \label{list:ss-1}
    \item Apply a compensation method for the zero loss phenomenon (default: change target labels to the second most likely classes) \label{list:ss-2}
    \item Apply a compensation method for the non-differentiability phenomenon (default: BPDA with softplus as a substitute for ReLU and $L_p$ norm pooling with $p=5$ for max pooling) \label{list:ss-3}
    \item Apply both compensation methods in \ref{list:ss-2} and \ref{list:ss-3} together \label{list:ss-4}
\end{enumerate}
This scheme results in 4 effective evaluations, and baseline attacks with stochasticity (e.g. R-FGSM) are set to have 4 random starts (i.e., a sample has to survive all four attempts to be considered accurate; in other words, this is equivalent to 4 cascading attacks but without compensation methods) for fair comparison. For iterative attacks, such as PGD, we cascade as:
\begin{enumerate}
    \item Apply an attack without any compensation methods (e.g., plain PGD). \label{list:iter-1}
    \item Apply PGD with an initialization method proposed in Sec 3.3 (default: PGD + Eigen) \label{list:iter-2}
    \item Apply a compensation method for the zero loss phenomenon with plain PGD (default: change target labels to the second most likely classes) \label{list:iter-3}
    \item Apply both compensation methods in \ref{list:iter-2} and \ref{list:iter-3} together \label{list:iter-4}
    \item Apply a compensation method for the non-differentiability phenomenon along with compensation methods in \ref{list:iter-2} and \ref{list:iter-3} (default: BPDA with softplus as a substitute for ReLU and $L_p$ norm pooling with $p=5$ for max pooling) \label{list:iter-5}
\end{enumerate}
Note that for iterative attacks, we do not test for every possible combination of compensation methods for the three phenomena. Resulting scheme has 5 effective evaluations, and baseline attacks are set to have 5 random starts. For PGD attacks in subsequent experiments, we use total 9 back-propagations (9 iterations without initialization methods of Sec 3.3 or 7 iterations with those initialization methods) as a default value. When a compensation method is not effective for a given model, just using another random start can be more effective than using that compensation method. In such case, accuracy against baseline attacks can be lower than accuracy against compensated attacks. 

%% file: app-writing/c-1.tex
In this section, we present additional results on benchmarking the trade-off between adversarial accuracy and model widths. Details of training hyperparameters for models used in this section are described in Table \ref{table:c-1-1}. 

\subfile{app-writing/table-c-1-1.tex}

\subfile{app-writing/fig-c-1-1.tex}
\textbf{Additional figures for CIFAR-10 dataset} We measure how accuracy difference among Simple models with different width is affected by the perturbation size $\epsilon$ of attacks (Fig \ref{fig:c-1-1} (a, b)). For both FG(S)M and PGD attacks, we observe that attacks with larger $\epsilon$ are affected more by compensation methods; in other words, accuracy against attacks with larger $\epsilon$ tends to be more overestimated for Simple models and CIFAR-10 dataset we consider in this example. 

We also measure adversarial accuracy for other architectures, Simple-BN models (Fig \ref{fig:c-1-1} (c)) and WRN 28 models (Fig \ref{fig:c-1-1} (d)). Similar to Simple models, the difference in adversarial accuracy between models with small and large widths is overestimated, and that difference reduces when compensation methods are applied. 

\subfile{app-writing/fig-c-1-2.tex}
\textbf{SVHN} For SVHN dataset, we consider Simple-BN and WRN 28 models (same architectures as those for CIFAR-10 dataset) (Fig \ref{fig:c-1-2}). In contrast to models trained on CIFAR-10, models trained on SVHN show less accuracy difference even for baseline attacks. Interestingly, adversarial accuracy of larger models often drops (e.g. accuracy against FGSM attacks for Simple-BN with width=16), and compensating can increase accuracy difference between small and large models in such case. Nevertheless, general magnitude of accuracy difference between small and large models is moderate ($\sim6\%$) compared to that for models trained on CIFAR-10 ($\sim25\%$). Results on SVHN indicate that even the same architectures and training conditions can show different behaviors depending on datasets.

\subfile{app-writing/fig-c-1-3.tex}
\textbf{TinyImageNet} We consider VGG 11, VGG-BN 11, and WRN 50 models with different relative widths for TinyImageNet (Fig \ref{fig:c-1-3}). VGG 11 and VGG-BN 11 models show similar pattern as models trained on CIFAR-10; adversarial accuracy of models with larger widths is overestimated, resulting in large accuracy difference between models for baseline attacks. However, behavior of WRN 50 models is more similar to that of models trained on SVHN, in that there is not significant difference in accuracy among models with different width. 

%% file: app-writing/table-c-1-1.tex
\begin{table}[h]
\centering
\caption{Training hyperparamters of models used in Appendix C.1.}
\vspace{0.5em}
\small
\begin{tabular}{@{}c|c|l@{}}
\toprule
Dataset                                                                   & Architecture                                                & Training condition                                                                                                                                                                                                                                         \\ \midrule
\multirow{2}{*}{\begin{tabular}[c]{@{}c@{}}CIFAR-10,\\ SVHN\end{tabular}} & Simple                                                      & \begin{tabular}[c]{@{}l@{}}Epochs: 100, Batch size: 128\\ Optimizer: SGD with momentum of 0.9\\ Learning rate: start with 0.01, decay by factor of 10 every 40 epochs\\ Regularization: fixed weight decay of $5\times10^{-4}$\end{tabular}                \\ \cmidrule(l){2-3} 
                                                                          & \begin{tabular}[c]{@{}c@{}}Simple-BN,\\ WRN 28\end{tabular} & \begin{tabular}[c]{@{}l@{}}Epochs: 100, Batch size: 128\\ Optimizer: SGD with momentum of 0.9\\ Learning rate: start with 0.1, decay by factor of 10 every 40 epochs\\ Regularization: fixed weight decay of $5\times10^{-4}$\end{tabular}                 \\ \midrule
\multirow{3}{*}{TinyImageNet}                                             & VGG 11                                                      & \begin{tabular}[c]{@{}l@{}}Epochs: 100, Batch size: 128 (96 for scale factor 4)\\ Optimizer: Adam ($\beta_1=0.9, \beta_2=0.99$)\\ Learning rate: start with $10^{-4}$. decay by factor of 10 every 40 epochs\\ Regularization: none\end{tabular}           \\ \cmidrule(l){2-3} 
                                                                          & VGG-BN 11                                                   & \begin{tabular}[c]{@{}l@{}}Epochs: 100, Batch size: 128 (96 for scale factor 4)\\ Optimizer: Adam ($\beta_1=0.9, \beta_2=0.99$)\\ Learning rate: start with $10^{-3}$. decay by factor of 10 every 40 epochs\\ Regularization: none\end{tabular}           \\ \cmidrule(l){2-3} 
                                                                          & WRN 50                                                      & \begin{tabular}[c]{@{}l@{}}Epochs: 100, Batch size: 128\\ Optimizer: Adam ($\beta_1=0.9, \beta_2=0.99$)\\ Learning rate: start with $10^{-3}$. decay by factor of 10 every 40 epochs\\ Regularization: fixed weight decay of $5\times10^{-4}$\end{tabular} \\ \bottomrule
\end{tabular}
\label{table:c-1-1}
\end{table}

%% file: app-writing/fig-c-1-1.tex
\begin{figure}[h]
    \centering
    \subfigure[]{\includegraphics[width=0.4\textwidth]{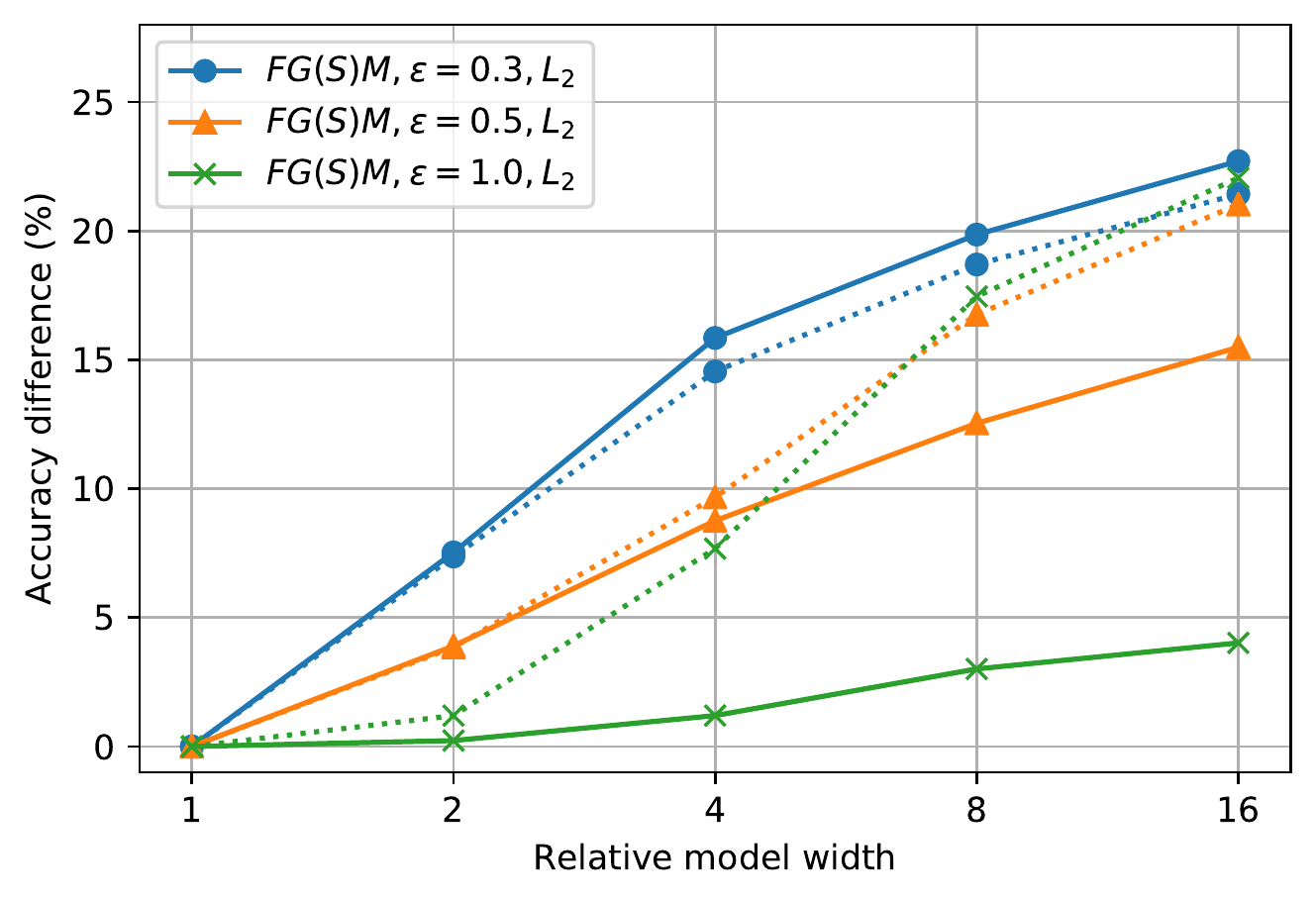}} \quad
    \subfigure[]{\includegraphics[width=0.4\textwidth]{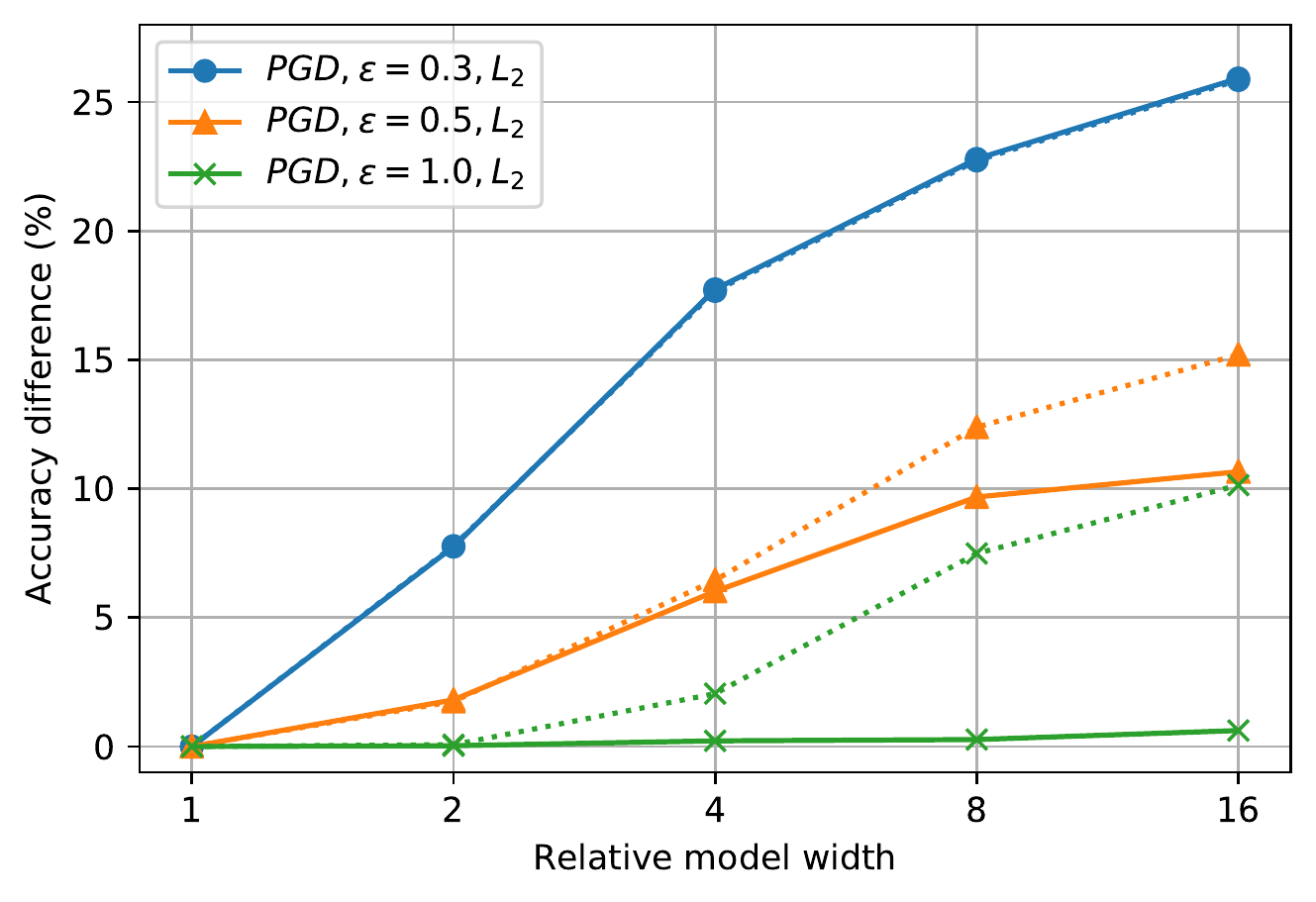}} \quad
    \subfigure[]{\includegraphics[width=0.4\textwidth]{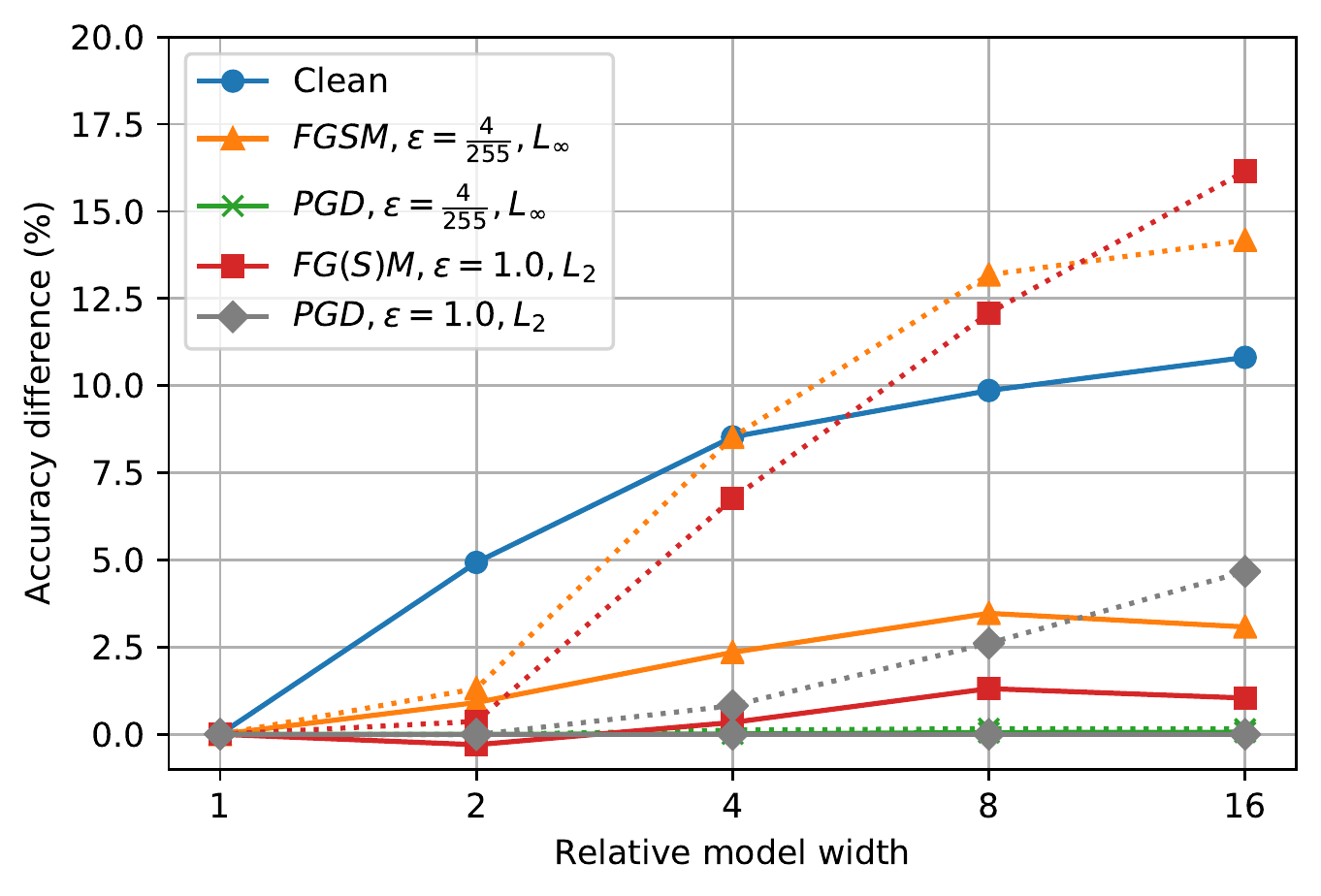}} \quad
    \subfigure[]{\includegraphics[width=0.4\textwidth]{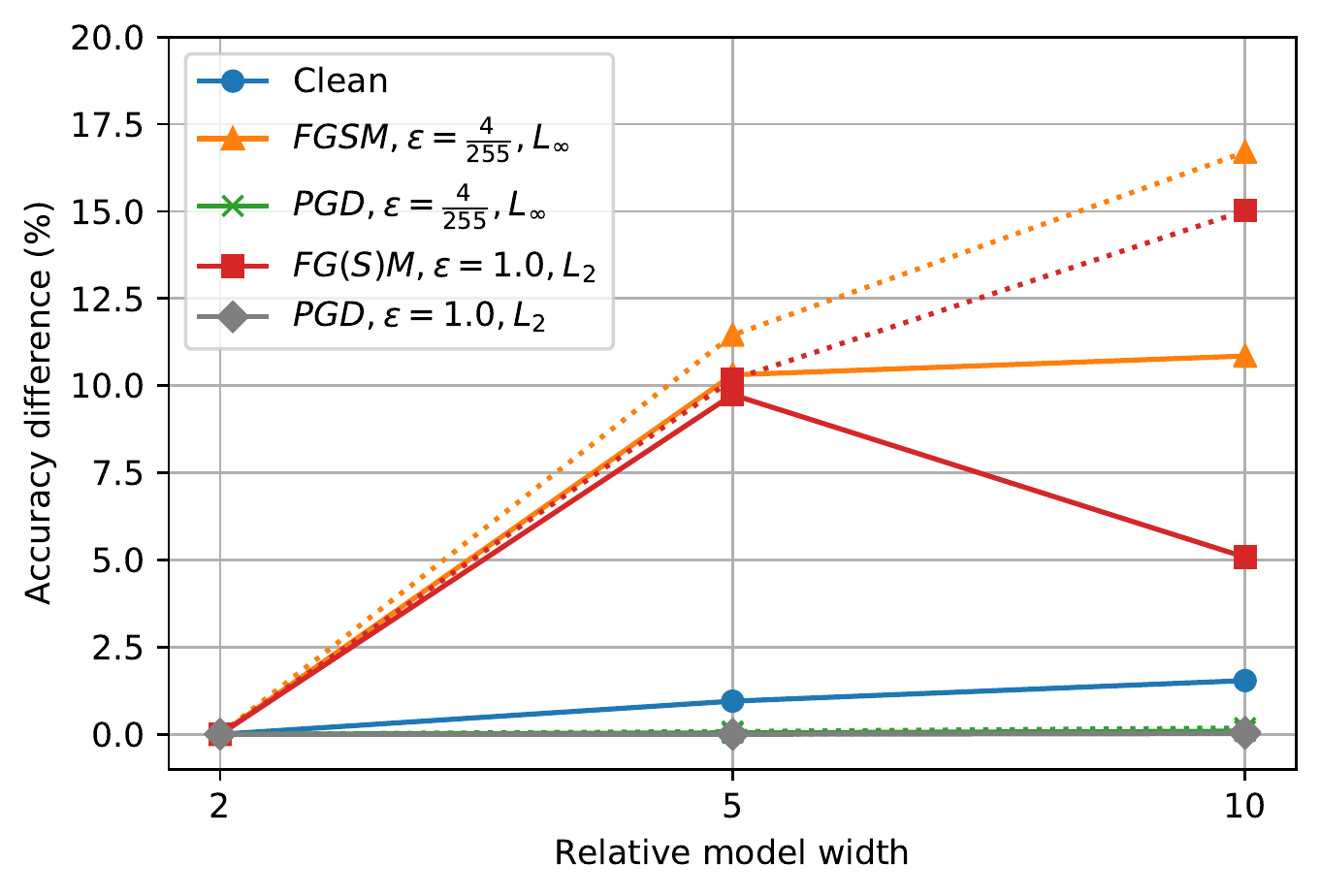}}
    \caption{(a, b) Accuracy difference (with respect to the accuracy of the model with width=1) of Simple models with different relative widths, measured for different perturbation sizes $\epsilon$ in $L_2$ norm using a single-step FG(S)M attack (a) and an iterative PGD attack (b). (c, d) Accuracy difference (with respect to the accuracy of the model with width=1 for Simple-BN and width=2 for WRN 28) of Simple-BN models (c) and WRN 28 models (d) with different relative widths. For all plots, dashed and solid lines represent accuracy against baseline and compensated attacks, respectively.}
    \label{fig:c-1-1}
\end{figure}

%% file: app-writing/fig-c-1-2.tex
\begin{figure}[h]
    \centering
    \subfigure[]{\includegraphics[width=0.4\textwidth]{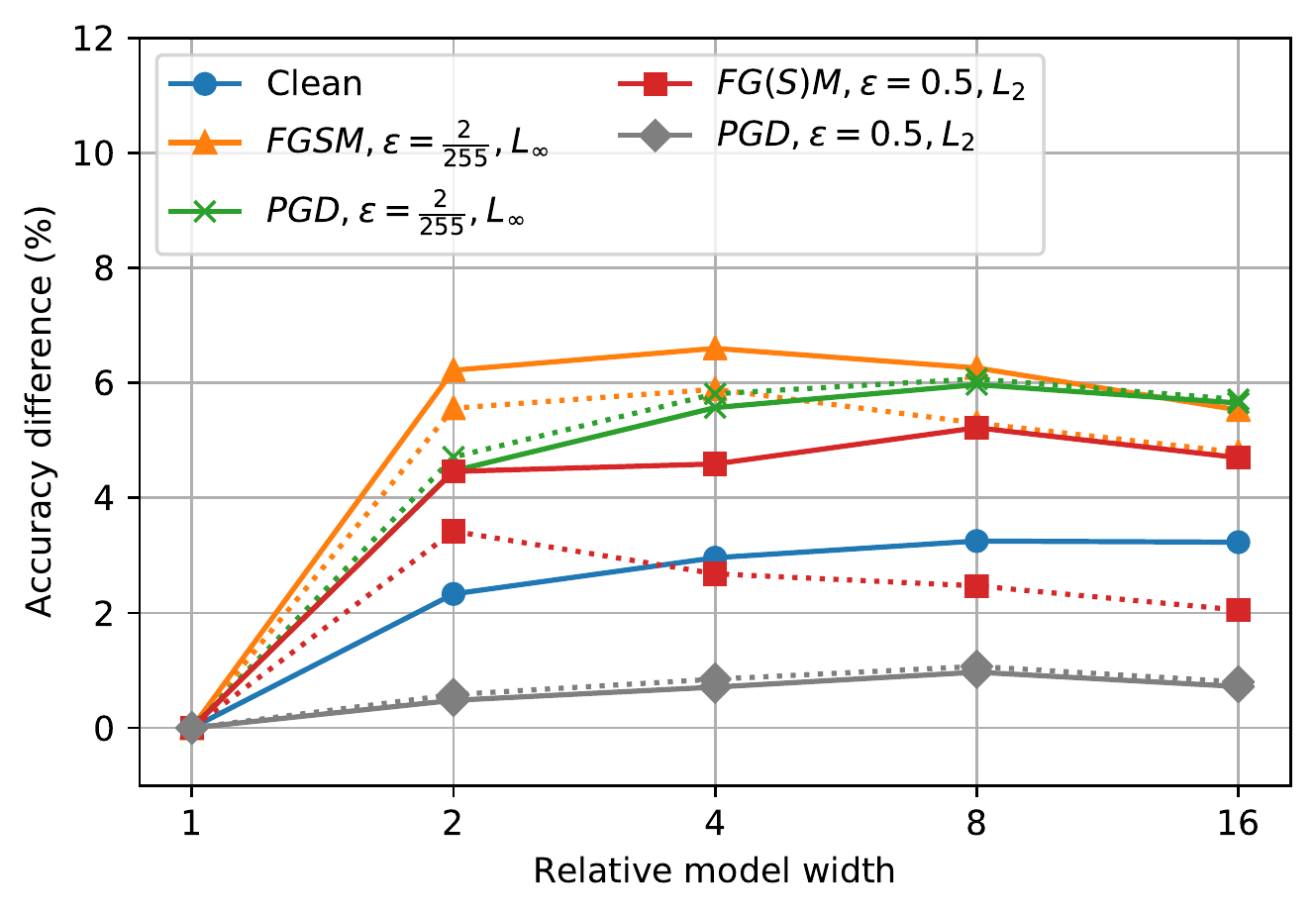}} \quad
    \subfigure[]{\includegraphics[width=0.4\textwidth]{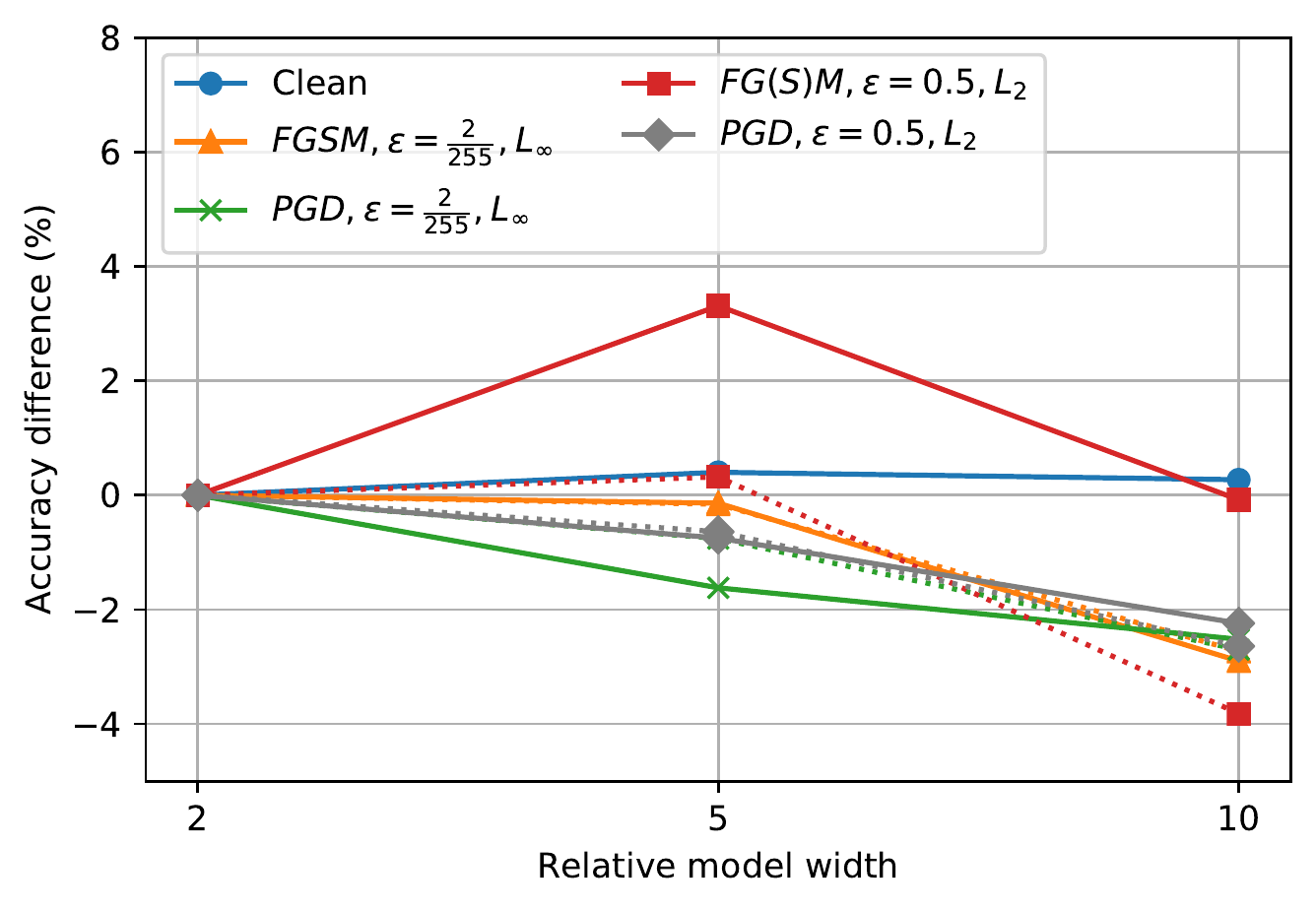}}
    \caption{Accuracy difference (with respect to the accuracy of the model with width=1 for Simple-BN and width=2 for WRN 28) of Simple-BN models (a) and WRN 28 models (b) trained on SVHN with different relative widths. Dashed and solid lines represent accuracy against baseline and compensated attacks, respectively.}
    \label{fig:c-1-2}
\end{figure}

%% file: app-writing/fig-c-1-3.tex
\begin{figure}[h]
    \centering
    \subfigure[]{\includegraphics[width=0.3\textwidth]{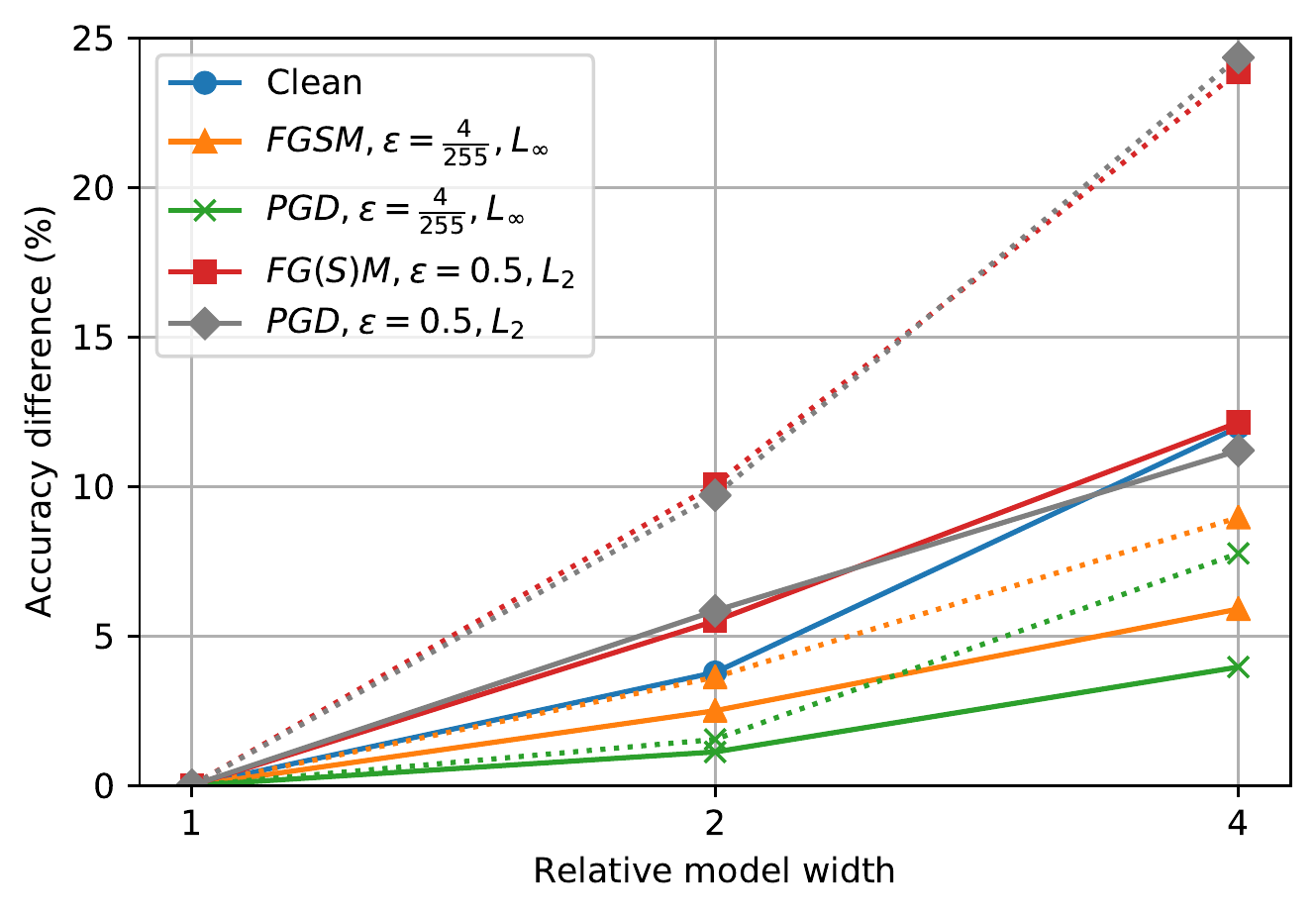}} 
    \subfigure[]{\includegraphics[width=0.3\textwidth]{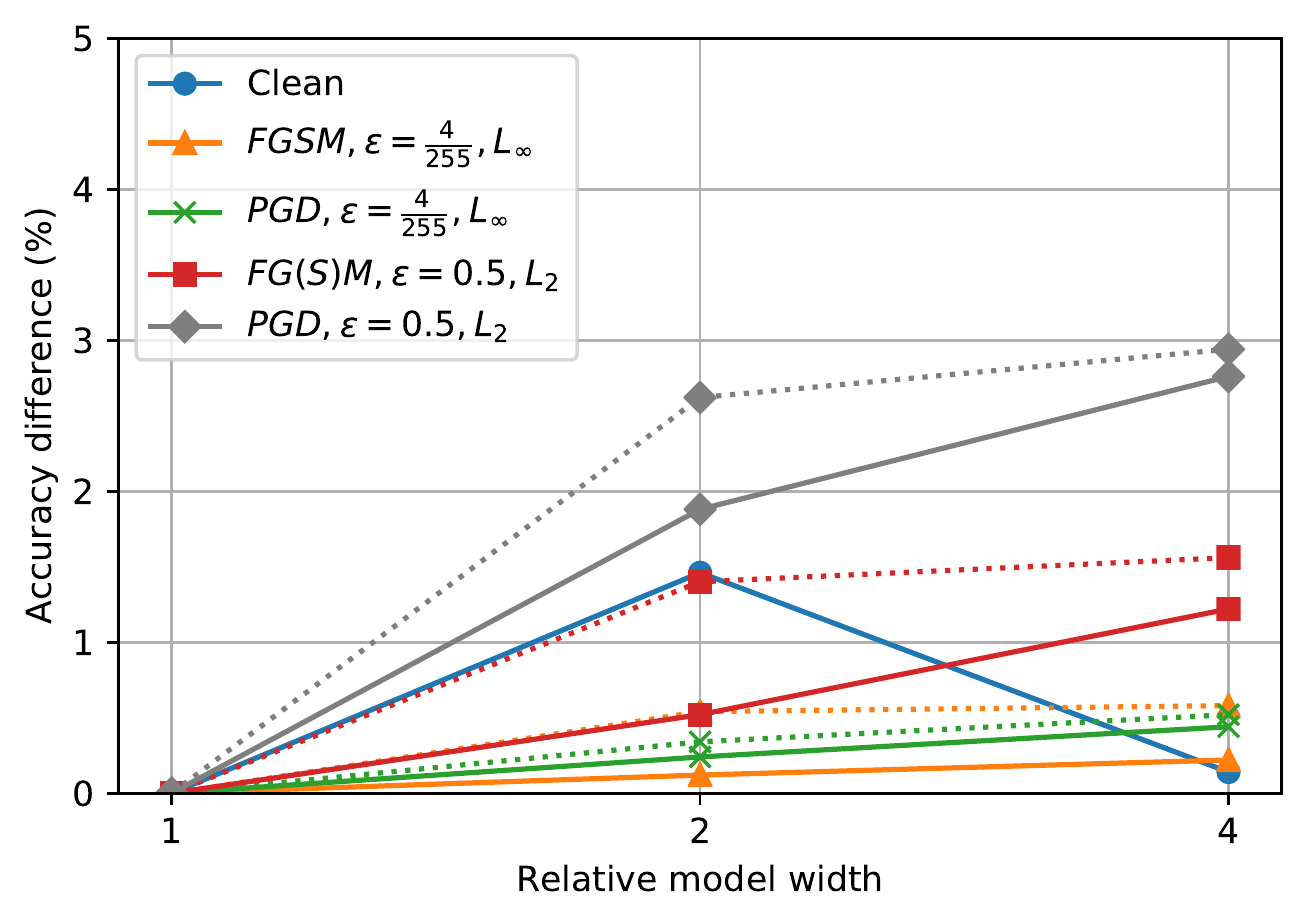}}
    \subfigure[]{\includegraphics[width=0.3\textwidth]{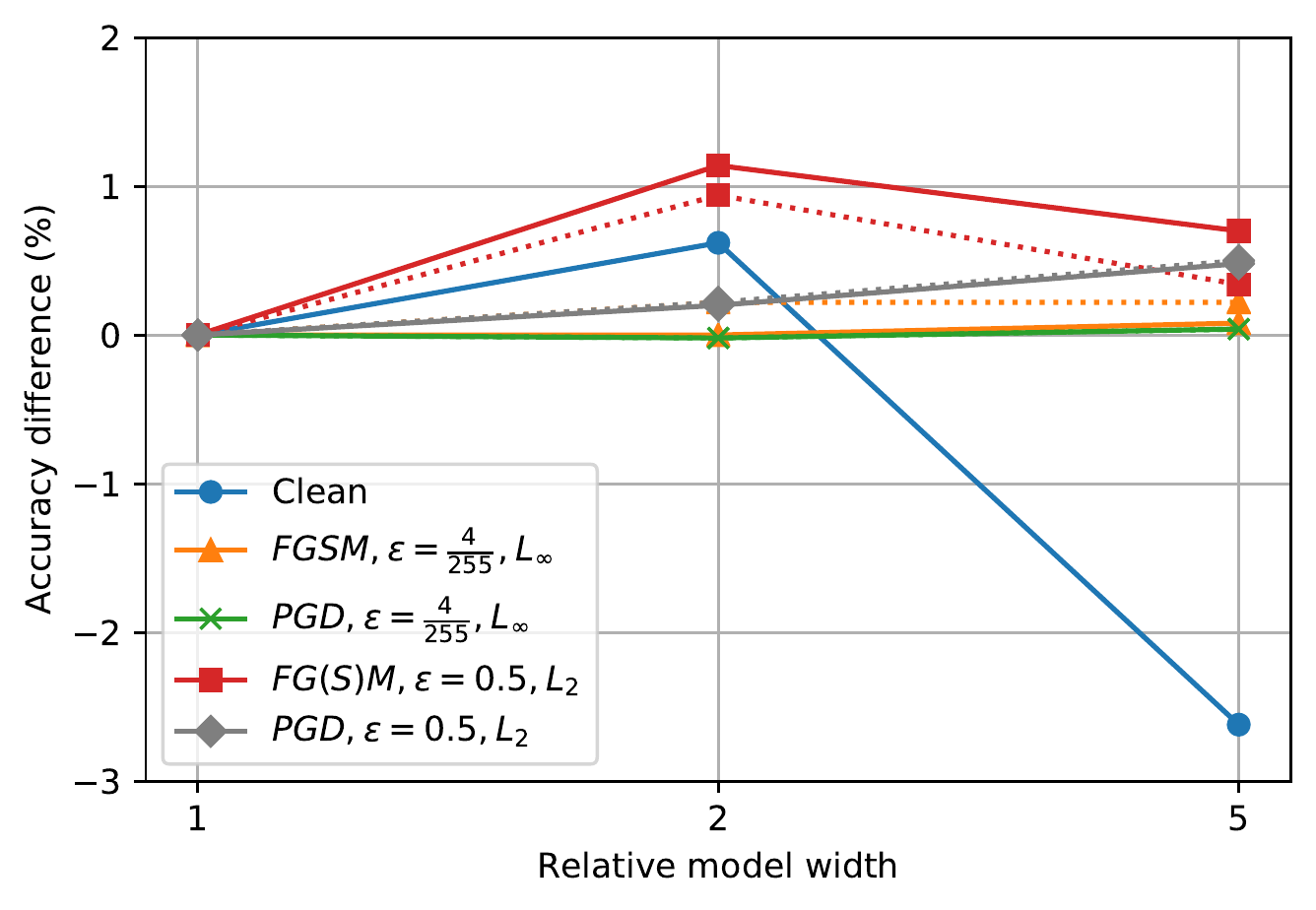}}
    \caption{Accuracy difference (with respect to the model with width=1) of VGG 11 models (a), VGG-BN 11 models (b), and WRN 50 models (c) trained on TinyImageNet with different relative widths. Dashed and solid lines represent accuracy against baseline and compensated attacks, respectively.}
    \label{fig:c-1-3}
\end{figure}

%% file: app-writing/c-2.tex
For weight pruning, we initially train a WRN 28 model with width scale factor of 10. The model is trained for 100 epochs using SGD with momentum of 0.9 as an optimizer, with starting learning rate of 0.1, which is decayed by factor of 10 every 40 epochs. We use early stopping based on the validation accuracy. Batch size is fixed to 128. To compare the impact of using weight decay, we train two models with and without weight decay of $5\times10^{-4}$ with otherwise same training conditions as stated. 

We iteratively remove weights with small magnitude as in typical weight pruning. To be specific, in each pruning iteration, we remove 25\% of total weights from convolution layers based on their magnitude, and finetune for 10 epochs with learning rate of 0.001. Otherwise, finetuning conditions are same as training conditions. However, note that the optimizer is initialized at each pruning iteration so that the momentum term from previous iteration (which contains information on removed weights) does not affect current finetuning. We iterate this process for 9 cycles, and the proportion of surviving weights at the final stage is 7.5\%. 

%% file: app-writing/c-3.tex
Regularization techniques are developed for better generalization, as for weight decay and spectral normalization \cite{miyato2018spectral}, or robustness against perturbations, as for orthonormal regularization \cite{cisse2017parseval}, Jacobian regularization \cite{ross2017improving, jakubovitz2018improving}, and adversarial training \cite{goodfellow2014explaining, madry2017towards}. In this section, we present these techniques in detail, and provide additional experimental results. 

%% file: app-writing/c-3-1.tex
Weight decay penalizes the Frobenius norm of weight matrices, inducing weights to be smaller. Typically, weight decay is directly incorporated to optimizers. However, when written as an additional regularization term to loss function, weight decay can be expressed as:
\begin{gather}
    l^\prime(f, x, t) = l(f(x), t) + \lambda\sum_{i=1}^{L} \|W_i\|_F^2
\end{gather}
where $f$ is a deep neural network model with total $L$ layers, $x$ is an input sample, and $t$ is the ground truth label. $l(\cdot, \cdot)$ is a standard cross-entropy loss, and $\lambda$ controls the strength of weight decay term. In this work, we consider fixed $\lambda$ throughout training process and increasing $\lambda$ as training progresses to excessively apply weight decay. $W_i$ represents a weight matrix of the $i$th layer. Typically, we use $\lambda=5\times10^{-4}$ for fixed weight decay, and start with $\lambda=1\times10^{-4}$ and multiply it by factor of 10 every 40 epochs for excessive weight decay. We choose to increase $\lambda$ instead of using large fixed $\lambda$ to study excessive weight decay, because larger $\lambda$ (e.g. $10^{-2}$) at earlier epochs resulted in poor training that often did not escape starting loss and accuracy level. 

Spectral normalization \cite{miyato2018spectral} was proposed to set the Lipschitz constant of each layer to be 1, by dividing weights of each layer with the estimated largest eigenvalue of that layer, so that training can be stabilized especially for Generative Adversarial Networks. For a weight matrix $W_i$ corresponding to the $i$th layer, \citeauthor{miyato2018spectral} (\citeyear{miyato2018spectral}) computed the largest singular value of this layer $\sigma(W_i)$ using a single-step power iteration per a single forward-backward pass during training, then divide the weight matrix by $\sigma(W_i)$ to set the large singular value to be 1. We use spectral normalization layer after each convolution and linear layer (except for the final linear classifier), and remove batch normalization if it is originally used. Also, we do not use spectral normalization for convolution layers used for residual connections in WRN 28. 

Orthonormal regularization was used as a part of Parseval network \cite{cisse2017parseval} that improved adversarial robustness of deep neural networks. In particular, orthonormal regularization induces each weight matrix to be orthonormal, so that the eigenvalues of $W_i^T W_i$ become 1. The motivation behind this is similar to that of spectral normalization, in that both aim to control the Lipschitz constant of each layer. The resulting loss function to be optimized is:
\begin{gather}
    l^\prime(f, x, t) = l(f(x), t) + \lambda\sum_{i=1}^{L}\|W_i^T W_i - I\|_F^2
\end{gather}
where $I$ is an identity matrix with the same size as $W_i^T W_i$. Note that this formulation is simplified from what \citeauthor{cisse2017parseval} (\citeyear{cisse2017parseval}) used in their work, which included sampling and other regularization for convexity. We follow orthonormal regularization used as in \citeauthor{lin2018defensive} (\citeyear{lin2018defensive}) that adopted this simple formulation for convolution layers to improve adversarial robustness of models with activation quantization. 

Jacobian regularization \cite{jakubovitz2018improving} or input gradient regularization \cite{ross2017improving} can be thought as reducing the first-order term in Taylor's expansion when a small perturbation $r$ is added to an input $x$:
\begin{gather}
    l(f(x+r), t) \simeq l(f(x), t) + r^T\cdot\nabla_x l(f(x), t) + O(\|r\|^2)
\end{gather}
Jacobian regularization computes gradients for every logit $z_i$ (where $z=f(x)$) with respect to the input $x$, and the resulting loss function is:
\begin{gather}
    l^\prime(f, x, t) = l(f(x), t) + \lambda\sqrt{\sum_{i=1}^{C}\sum_{j=1}^{d_x} (\frac{\partial z_i}{\partial x_j})^2}
\end{gather}
This method can require significant additional back-propagations when the total number of classes $C$ is large, as in the case of TinyImageNet with 200 classes. More simply, input gradient regularization computes Jacobian not from logits $z$, but from loss directly:
\begin{gather}
    l^\prime(f, x, t) = l(f(x), t) + \lambda\sum_{j=1}^{d_x} (\frac{\partial l(f(x), t)}{\partial x_j})^2
\end{gather}
In this work, we consider the later form of input gradient regularization. For more details including theoretical justification of Jacobian regularization in relation to Lipschitz stability, we refer readers to \citeauthor{jakubovitz2018improving} (\citeyear{jakubovitz2018improving}).

Adversarial training \cite{goodfellow2014explaining, madry2017towards} achieved robustness by directly train a model on samples crafted by attack methods. \citeauthor{madry2017towards} (\citeyear{madry2017towards}) analyzed that adversarial training can be thought as an optimization for min-max problem when a model $f$ is parameterized with $\theta$:
\begin{gather}
    \theta^* = \underset{\theta}{\arg\min} \ l(f_\theta(x_{adv}), t) = \underset{\theta}{\arg\min}\ \underset{x^\prime}{\max}\ l(f_\theta(x^\prime), t)
\end{gather}
and that attack methods, such as PGD, are approximation for the inner maximization.

%% file: app-writing/c-3-2.tex
In addition to the result presented in Section 4.2, we show how adversarial accuracy of regularization techniques is affected by the three phenomena, for different architectures and datasets. Details of training conditions for models used in this section are shown in Table \ref{table:c-3-1}.

\subfile{app-writing/table-c-3-1.tex}

First, we present more detailed results for WRN 28 models trained on CIFAR-10 using different regularization techniques in Table \ref{table:c-3-2}. We measure adversarial accuracy against both baseline and compensated attacks for different perturbation sizes $\epsilon$ in $L_\infty$ and $L_2$ norm. Generally, adversarial accuracy against smaller perturbation $\epsilon$ tends to be less affected by compensation methods, manifested by relatively small gap between baseline and compensated adversarial accuracy for $\epsilon=\frac{2}{255}$ for $L_\infty$ norm and $\epsilon=0.3$ for $L_2$ norm. 
\subfile{app-writing/table-c-3-2.tex}

Then, we consider different architectures, Simple and Simple-BN, for CIFAR-10 dataset (Table \ref{table:c-3-3}). We observe that spectral normalization and orthonormal regularization are less affected by compensation methods for these architectures, in contrast to WRN 28 architecture. For example, adversarial accuracy (PGD, $\epsilon=\frac{4}{255})$ of a WRN 28 with spectral normalization decreases by 5.25\% when compensation methods are applied, but that for a Simple model only decreases by 0.02\%. For a Simple model with orthonormal regularization, compensation methods with PGD turn out to be not effective and result in even higher adversarial accuracy compared to baseline PGD. Therefore, the effectiveness of regularization methods, especially whether they rely on the three phenomena inflating adversarial accuracy, might be dependent on model architectures. 
\subfile{app-writing/table-c-3-3.tex}

Finally, we run similar experiments for different datasets, SVHN (Table \ref{table:c-3-4}) and TinyImageNet (Table \ref{table:c-3-5}). Models trained on SVHN generally show much less gap between adversarial accuracy against baseline and compensated attacks compared to those trained on CIFAR-10, even when architectures are exactly same for both SVHN and CIFAR-10. Also, adversarial accuracy of models using weight decay and spectral normalization is affected less severely by compensation methods for SVHN. WRN 50 models trained on TinyImageNet are benchmarked for limited regularizaton techniques. In contrast to CIFAR-10 and SVHN, Jacobian regularization seems to be affected by compensation methods by $6\%$ of accuracy difference against a PGD attack. Overall, these observations imply that whether adversarial accuracy of regularizaton techniques is overestimated can be complicated by model architectures and datasets. 
\subfile{app-writing/table-c-3-4.tex}
\subfile{app-writing/table-c-3-5.tex}

%% file: app-writing/table-c-3-1.tex
\begin{table}[h]
\centering
\caption{Training hyperparameters of models used in Appendix C.3}
\vspace{0.5em}
\small
\begin{tabular}{|c|c|l|l|}
\hline
Dataset                       & Architecture                                                                   & \begin{tabular}[c]{@{}l@{}}Regularization\\ ($\lambda$ if applicable)\end{tabular}                                                                                                                                                 & Other training conditions                                                                                                                                                                           \\ \hline
\multirow{20}{*}{CIFAR-10}    & \multirow{7}{*}{\begin{tabular}[c]{@{}c@{}}Simple\\ (width=4)\end{tabular}}    & None                                                                                                                                                                                                                               & \multirow{7}{*}{\begin{tabular}[c]{@{}l@{}}Epochs: 100,\\ Batch size: 128,\\ Optimizer: SGD (momentum=0.9),\\ Learning rate: start with 0.01, decay by 10 / 40 epochs\end{tabular}}                 \\ \cline{3-3}
                              &                                                                                & Weight decay, $\lambda=5\times10^{-4}$                                                                                                                                                                                             &                                                                                                                                                                                                     \\ \cline{3-3}
                              &                                                                                & Weight decay excess                                                                                                                                                                                                                &                                                                                                                                                                                                     \\ \cline{3-3}
                              &                                                                                & Spectral normalization                                                                                                                                                                                                             &                                                                                                                                                                                                     \\ \cline{3-3}
                              &                                                                                & Orthonormal, $\lambda=1\times10^{-4}$                                                                                                                                                                                              &                                                                                                                                                                                                     \\ \cline{3-3}
                              &                                                                                & Input gradient, $\lambda=1.0$                                                                                                                                                                                           &                                                                                                                                                                                                     \\ \cline{3-3}
                              &                                                                                & \begin{tabular}[c]{@{}l@{}}Adversarial training,\\ for $L_\infty$ norm: $\epsilon=\frac{8}{255}$, 7 iterations\\ for $L_2$ norm: $\epsilon=1.0$, 7 iterations \\ Apply fixed weight decay of $\lambda=5\times10^{-4}$\end{tabular} &                                                                                                                                                                                                     \\ \cline{2-4} 
                              & \multirow{6}{*}{\begin{tabular}[c]{@{}c@{}}Simple-BN\\ (width=4)\end{tabular}} & None                                                                                                                                                                                                                               & \multirow{13}{*}{\begin{tabular}[c]{@{}l@{}}Learning rate: start with 0.1\\ Otherwise same as above\end{tabular}}                                                                                   \\ \cline{3-3}
                              &                                                                                & Weight decay, $\lambda=5\times10^{-4}$                                                                                                                                                                                             &                                                                                                                                                                                                     \\ \cline{3-3}
                              &                                                                                & Weight decay excess                                                                                                                                                                                                                &                                                                                                                                                                                                     \\ \cline{3-3}
                              &                                                                                & Orthonormal, $\lambda=1\times10^{-3}$                                                                                                                                                                                              &                                                                                                                                                                                                     \\ \cline{3-3}
                              &                                                                                & Input gradient, $\lambda=1.0$                                                                                                                                                                                           &                                                                                                                                                                                                     \\ \cline{3-3}
                              &                                                                                & \begin{tabular}[c]{@{}l@{}}Adversarial training,\\ for $L_\infty$ norm: $\epsilon=\frac{8}{255}$, 7 iterations\\ for $L_2$ norm: $\epsilon=1.0$, 7 iterations\\ Apply fixed weight decay of $\lambda=5\times10^{-4}$\end{tabular}  &                                                                                                                                                                                                     \\ \cline{2-3}
                              & \multirow{7}{*}{\begin{tabular}[c]{@{}c@{}}WRN 28\\ (width=2)\end{tabular}}    & None                                                                                                                                                                                                                               &                                                                                                                                                                                                     \\ \cline{3-3}
                              &                                                                                & Weight decay, $\lambda=5\times10^{-4}$                                                                                                                                                                                             &                                                                                                                                                                                                     \\ \cline{3-3}
                              &                                                                                & Weight decay excess                                                                                                                                                                                                                &                                                                                                                                                                                                     \\ \cline{3-3}
                              &                                                                                & Spectral normalization                                                                                                                                                                                                             &                                                                                                                                                                                                     \\ \cline{3-3}
                              &                                                                                & Orthonormal, $\lambda=1\times10^{-3}$                                                                                                                                                                                              &                                                                                                                                                                                                     \\ \cline{3-3}
                              &                                                                                & Input gradient, $\lambda=1.0$                                                                                                                                                                                           &                                                                                                                                                                                                     \\ \cline{3-3}
                              &                                                                                & \begin{tabular}[c]{@{}l@{}}Adversarial training,\\ for $L_\infty$ norm: $\epsilon=\frac{8}{255}$, 7 iterations\\ for $L_2$ norm: $\epsilon=1.0$, 7 iterations\\ Apply fixed weight decay of $\lambda=5\times10^{-4}$\end{tabular}  &                                                                                                                                                                                                     \\ \hline
\multirow{14}{*}{SVHN}        & \multirow{7}{*}{\begin{tabular}[c]{@{}c@{}}Simple-BN\\ (width=4)\end{tabular}} & None                                                                                                                                                                                                                               & \multirow{3}{*}{\begin{tabular}[c]{@{}l@{}}Epochs: 100,\\ Batch size: 128,\\ Optimizer: SGD (momentum=0.9),\\ Learning rate: start with 0.1, decay by 10 / 40 epochs\end{tabular}}                  \\ \cline{3-3}
                              &                                                                                & Weight decay, $\lambda=5\times10^{-4}$                                                                                                                                                                                             &                                                                                                                                                                                                     \\ \cline{3-3}
                              &                                                                                & \begin{tabular}[c]{@{}l@{}}Weight decay excess \vspace{1em} \end{tabular}                                                                                                                                                                                                                &                                                                                                                                                                                                     \\ \cline{3-4} 
                              &                                                                                & Spectral normalization                                                                                                                                                                                                             & \begin{tabular}[c]{@{}l@{}}Optimizer: Adam ($\beta_1=0.9, \beta_2=0.99$)\\ Learning rate: start with $10^{-4}$\\ Otherwise same as above.\end{tabular}                                              \\ \cline{3-4} 
                              &                                                                                & Orthonormal, $\lambda=1\times10^{-3}$                                                                                                                                                                                              & \multirow{2}{*}{Same as those for `None' (see above)}                                                                                                                                               \\ \cline{3-3}
                              &                                                                                & Input gradient, $\lambda=1.0$                                                                                                                                                                                           &                                                                                                                                                                                                     \\ \cline{3-4} 
                              &                                                                                & \begin{tabular}[c]{@{}l@{}}Adversarial training,\\ for $L_\infty$ norm: $\epsilon=\frac{8}{255}$, 7 iterations\\ for $L_2$ norm: $\epsilon=1.0$, 7 iterations\\ Apply fixed weight decay of $\lambda=5\times10^{-4}$\end{tabular}  & \begin{tabular}[c]{@{}l@{}}Optimizer: Adam ($\beta_1=0.5, \beta_2=0.5$)\\ Learning rate: start with $10^{-3}$\\ Otherwise same as above.\end{tabular}                                               \\ \cline{2-4} 
                              & \multirow{7}{*}{\begin{tabular}[c]{@{}c@{}}WRN 28\\ (width=2)\end{tabular}}    & None                                                                                                                                                                                                                               & \multirow{6}{*}{\begin{tabular}[c]{@{}l@{}}Epochs: 100,\\ Batch size: 128,\\ Optimizer: SGD (momentum=0.9),\\ Learning rate: start with 0.1, decay by 10 / 40 epochs\end{tabular}}                  \\ \cline{3-3}
                              &                                                                                & Weight decay, $\lambda=5\times10^{-4}$                                                                                                                                                                                             &                                                                                                                                                                                                     \\ \cline{3-3}
                              &                                                                                & Weight decay excess                                                                                                                                                                                                                &                                                                                                                                                                                                     \\ \cline{3-3}
                              &                                                                                & Spectral normalization                                                                                                                                                                                                             &                                                                                                                                                                                                     \\ \cline{3-3}
                              &                                                                                & Orthonormal, $\lambda=1\times10^{-3}$                                                                                                                                                                                              &                                                                                                                                                                                                     \\ \cline{3-3}
                              &                                                                                & Input gradient, $\lambda=1.0$                                                                                                                                                                                           &                                                                                                                                                                                                     \\ \cline{3-4} 
                              &                                                                                & \begin{tabular}[c]{@{}l@{}}Adversarial training,\\ for $L_\infty$ norm: $\epsilon=\frac{8}{255}$, 7 iterations\\ for $L_2$ norm: $\epsilon=1.0$, 7 iterations\\ Apply fixed weight decay of $\lambda=5\times10^{-4}$\end{tabular}  & \begin{tabular}[c]{@{}l@{}}Optimizer: Adam ($\beta_1=0.5, \beta_2=0.9$)\\ Learning rate: start with $10^{-3}$\\ Otherwise same as above.\end{tabular}                                               \\ \hline
\multirow{5}{*}{TinyImageNet} & \multirow{5}{*}{\begin{tabular}[c]{@{}c@{}}WRN 50\\ (width=1)\end{tabular}}    & None                                                                                                                                                                                                                               & \multirow{3}{*}{\begin{tabular}[c]{@{}l@{}}Epochs: 100,\\ Batch size: 128, \\ Optimizer: Adam ($\beta_1=0.9, \beta_2=0.99$),\\ Learning rate: start with 0.1, decay by 10 / 40 epochs\end{tabular}} \\ \cline{3-3}
                              &                                                                                & Weight decay, $\lambda=5\times10^{-4}$                                                                                                                                                                                             &                                                                                                                                                                                                     \\ \cline{3-3}
                              &                                                                                & \begin{tabular}[c]{@{}l@{}}Weight decay excess \vspace{1em} \end{tabular}                                                                                                                                                                                                                &                                                                                                                                                                                                     \\ \cline{3-4} 
                              &                                                                                & Input gradient, $\lambda=0.01$                                                                                                                                                                                          & \multirow{2}{*}{\begin{tabular}[c]{@{}l@{}}Batch size: 64\\ Otherwise same as above.\end{tabular}}                                                                                                  \\ \cline{3-3}
                              &                                                                                & Input gradient, $\lambda=0.05$                                                                                                                                                                                          &                                                                                                                                                                                                     \\ \hline
\end{tabular}
\label{table:c-3-1}
\end{table}

%% file: app-writing/table-c-3-2.tex
\begin{sidewaystable}[h]
\centering
\caption{Accuracy of WRN 28 models trained on CIFAR-10 using different regularization techniques against first-order attack methods in the order of FGSM/R-FGSM/PGD for stated perturbation sizes $\epsilon$ in $L_\infty$ norm (above) and $L_2$ norm (below). We compare adversarial accuracy against baseline and compensated attacks, as in Table 2 of Sec 4.2, which is a subset of this result.}
\vspace{0.5em}
\small
\begin{tabular}{@{}cccccccc@{}}
\toprule
\multirow{2}{*}{Regularization} & \multirow{2}{*}{Clean} & \multicolumn{2}{c}{$\epsilon=\frac{2}{255}$}  & \multicolumn{2}{c}{$\epsilon=\frac{4}{255}$}  & \multicolumn{2}{c}{$\epsilon=\frac{8}{255}$}  \\ \cmidrule(l){3-8} 
                                &                        & Baseline              & Compensated           & Baseline              & Compensated           & Baseline              & Compensated           \\ \midrule
None                            & 91.65                  & 33.20 / 46.17 / 1.59  & 16.48 / 36.06 / 1.23  & 21.90 / 23.78 / 0.02  & 5.94 / 11.25 / 0      & 17.57 / 7.51 / 0      & 3.36 / 1.30 / 0       \\
Weight decay                    & 93.64                  & 40.27 / 49.01 / 2.24  & 35.50 / 47.26 / 2.00  & 31.47 / 29.69 / 0.02  & 20.31 / 27.32 / 0.02  & 24.78 / 13.62 / 0     & 6.62 / 9.24 / 0       \\
Weight decay excess             & 91.52                  & 46.45 / 53.60 / 8.68  & 46.45 / 53.60 / 5.12  & 47.83 / 45.48 / 1.85  & 36.46 / 41.58 / 0.63  & 42.91 / 33.05 / 0.32  & 16.90 / 24.24 / 0.01  \\
Spectral norm                   & 87.31                  & 42.64 / 58.19 / 25.62 & 33.52 / 57.66 / 23.50 & 31.50 / 40.79 / 7.50  & 9.86 / 31.67 / 2.25   & 21.10 / 27.45 / 1.77  & 1.46 / 7.84 / 0.01    \\
Orthonormal                     & 93.47                  & 36.54 / 48.34 / 2.48  & 25.91 / 44.66 / 2.40  & 27.19 / 27.29 / 0.03  & 13.25 / 19.76 / 0.01  & 21.78 / 10.97 / 0     & 8.42 / 5.18 / 0       \\
Input gradient                        & 89.75                  & 55.06 / 74.21 / 49.14 & 53.52 / 73.72 / 48.93 & 24.93 / 53.03 / 12.80 & 23.18 / 52.14 / 12.79 & 7.51 / 20.72 / 0.33   & 3.56 / 19.86 / 0.25   \\
Adv Training                    & 82.67                  & 75.44 / 78.87 / 75.03 & 74.97 / 78.76 / 74.64 & 67.62 / 75.11 / 65.58 & 66.46 / 74.81 / 64.82 & 54.03 / 66.93 / 46.89 & 52.18 / 66.13 / 45.80 \\ \bottomrule
\end{tabular}
\newline
\vspace*{1em}
\newline
\begin{tabular}{@{}cccccccc@{}}
\toprule
\multirow{2}{*}{Regularization} & \multirow{2}{*}{Clean} & \multicolumn{2}{c}{$\epsilon=0.3$}            & \multicolumn{2}{c}{$\epsilon=0.5$}            & \multicolumn{2}{c}{$\epsilon=1.0$}            \\ \cmidrule(l){3-8} 
                                &                        & Baseline              & Compensated           & Baseline              & Compensated           & Baseline              & Compensated           \\ \midrule
None                            & 91.65                  & 49.86 / 53.78 / 26.29 & 19.59 / 38.64 / 0.83  & 47.30 / 47.11 / 21.47 & 11.72 / 21.93 / 0.01  & 45.33 / 39.07 / 15.20 & 7.30 / 7.73 / 0       \\
Weight decay                    & 93.64                  & 48.53 / 55.92 / 1.82  & 41.84 / 54.56 / 1.56  & 41.81 / 45.10 / 0.07  & 28.21 / 42.17 / 0.02  & 35.43 / 28.88 / 0     & 11.26 / 20.55 / 0     \\
Weight decay excess             & 91.52                  & 56.10 / 57.61 / 9.21  & 51.12 / 56.10 / 4.55  & 53.23 / 53.74 / 2.79  & 44.23 / 50.97 / 0.75  & 50.48 / 44.72 / 0.92  & 26.82 / 37.35 / 0.09  \\
Spectral norm                   & 87.31                  & 43.34 / 58.44 / 28.15 & 33.51 / 57.80 / 22.25 & 38.30 / 45.49 / 21.52 & 16.43 / 39.40 / 4.45  & 34.96 / 36.50 / 18.72 & 4.48 / 15.39 / 0.03   \\
Orthonormal                     & 93.47                  & 43.24 / 53.37 / 2.30  & 29.43 / 47.18 / 1.73  & 36.75 / 41.29 / 0.15  & 19.83 / 31.24 / 0.02  & 30.57 / 25.51 / 0.05  & 13.16 / 14.42 / 0     \\
Input gradient                        & 89.75                  & 54.07 / 74.04 / 48.41 & 52.46 / 73.48 / 48.10 & 32.67 / 60.09 / 19.55 & 30.69 / 59.20 / 19.51 & 10.32 / 30.43 / 1.90  & 6.38 / 29.31 / 1.16   \\
Adv Training                    & 83.07                  & 73.87 / 78.35 / 73.36 & 73.40 / 78.21 / 73.00 & 67.27 / 75.27 / 65.56 & 66.58 / 75.08 / 64.98 & 52.33 / 67.14 / 47.23 & 50.61 / 66.52 / 46.31 \\ \bottomrule
\end{tabular}

\label{table:c-3-2}
\end{sidewaystable}

%% file: app-writing/table-c-3-3.tex
\begin{table}[h]
\centering
\caption{Accuracy of Simple and Simple-BN models trained on CIFAR-10 using different regularization techniques against baseline and compensated attack methods, in the order of FGSM/R-FGSM/PGD with $\epsilon=\frac{4}{255}$ in $L_\infty$ norm. Note that spectral normalization for Simple-BN is \emph{same} as that for Simple, as spectral normalization layer is used instead of batch normalization layer.}
\vspace{0.5em}
\small
\begin{tabular}{@{}cccc|ccc@{}}
\toprule
\multirow{2}{*}{Regularization} & \multicolumn{3}{c|}{Simple}                           & \multicolumn{3}{c}{Simple-BN}                         \\ \cmidrule(l){2-7} 
                                & Clean & Baseline              & Compensated           & Clean & Baseline              & Compensated           \\ \midrule
None                            & 84.75 & 19.50 / 30.78 / 2.55  & 8.74 / 28.31 / 1.58   & 87.09 & 28.66 / 29.81 / 6.26  & 6.89 / 17.93 / 0.07   \\
Weight decay                    & 85.06 & 17.74 / 35.30 / 3.40  & 11.08 / 33.35 / 3.08  & 89.84 & 13.62 / 18.32 / 0.13  & 4.69 / 15.15 / 0.01   \\
Weight decay excess             & 84.92 & 15.14 / 36.30 / 3.29  & 11.64 / 34.22 / 3.20  & 87.58 & 6.10 / 16.80 / 0.03   & 4.90 / 15.33 / 0.03   \\
Spectral norm                   & 81.47 & 22.33 / 47.20 / 13.32 & 20.29 / 45.96 / 13.20 & -     & -                     & -                     \\
Orthonormal                     & 84.82 & 16.38 / 39.05 / 5.12  & 12.89 / 37.01 / 5.21  & 88.59 & 9.53 / 18.92 / 0.07   & 4.05 / 16.25 / 0.06   \\
Input gradient                        & 84.26 & 24.98 / 50.24 / 14.91 & 22.81 / 48.86 / 14.84 & 84.67 & 26.74 / 50.98 / 15.02 & 23.71 / 49.68 / 15.00 \\
Adv training                    & 67.04 & 54.14 / 60.61 / 53.48 & 52.95 / 60.05 / 52.19 & 70.91 & 57.40 / 64.07 / 56.48 & 56.36 / 63.54 / 55.43 \\ \bottomrule
\end{tabular}
\label{table:c-3-3}
\end{table}

%% file: app-writing/table-c-3-4.tex
\begin{table}[h]
\centering
\caption{Accuracy of Simple-BN and WRN 28 models trained on SVHN using different regularization techniques against baseline and compensated attack methods, in the order of FGSM/R-FGSM/PGD with $\epsilon=\frac{4}{255}$ in $L_\infty$ norm. }
\vspace{0.5em}
\small
\begin{tabular}{@{}cccc|ccc@{}}
\toprule
\multirow{2}{*}{Regularization} & \multicolumn{3}{c|}{Simple-BN}                        & \multicolumn{3}{c}{WRN 28}                            \\ \cmidrule(l){2-7} 
                                & Clean & Baseline              & Compensated           & Clean & Baseline              & Compensated           \\ \midrule
None                            & 94.29 & 28.60 / 45.39 / 2.80  & 19.10 / 41.31 / 2.38  & 95.42 & 49.74 / 60.31 / 4.06  & 34.30 / 55.05 / 3.65  \\
Weight decay                    & 95.37 & 28.06 / 51.61 / 4.57  & 23.90 / 49.64 / 4.64  & 96.38 & 63.99 / 70.25 / 8.71  & 56.25 / 68.12 / 7.31  \\
Weight decay excess             & 92.85 & 26.76 / 51.31 / 6.51  & 22.56 / 49.19 / 6.33  & 95.03 & 57.56 / 66.51 / 13.86 & 51.95 / 64.31 / 12.14 \\
Spectral norm                   & 88.26 & 32.83 / 59.93 / 22.83 & 29.45 / 57.95 / 22.50 & 94.63 & 40.91 / 63.94 / 22.05 & 37.03 / 61.84 / 21.97 \\
Orthonormal                     & 94.89 & 24.37 / 46.15 / 1.72  & 15.24 / 40.82 / 1.66  & 96.25 & 56.78 / 65.28 / 5.78  & 39.59 / 59.80 / 4.87  \\
Input gradient                        & 91.43 & 39.50 / 65.73 / 25.52 & 34.64 / 63.38 / 24.99 & 95.45 & 50.22 / 76.16 / 35.49 & 46.49 / 74.80 / 35.13 \\
Adv training                    & 83.75 & 66.63 / 74.87 / 63.41 & 64.07 / 74.34 / 61.95 & 91.91 & 78.27 / 85.15 / 75.58 & 76.16 / 84.76 / 74.47 \\ \bottomrule
\end{tabular}
\label{table:c-3-4}
\end{table}

%% file: app-writing/table-c-3-5.tex
\begin{table}[h]
\centering
\caption{Accuracy of WRN 50 models trained on TinyImageNet using different regularization techniques against baseline and compensated attack methods, in the order of FGSM/R-FGSM/PGD with $\epsilon=\frac{2}{255}$ in $L_\infty$ norm.}
\vspace{0.5em}
\small
\begin{tabular}{@{}cccc@{}}
\toprule
Regularization            & Clean & Baseline              & Compensated          \\ \midrule
None                      & 57.24 & 24.38 / 27.90 / 8.50  & 10.40 / 23.76 / 5.30 \\
Weight decay              & 55.22 & 5.80 / 14.22 / 0.68   & 4.68 / 13.30 / 0.64  \\
Weight decay excess       & 57.54 & 9.56 / 19.88 / 2.06   & 7.04 / 17.64 / 1.84  \\
Input gradient ($\lambda=0.01$) & 53.70 & 24.66 / 27.96 / 12.94 & 10.76 / 24.88 / 6.68 \\
Input gradient ($\lambda=0.05$) & 55.58 & 24.56 / 28.62 / 14.52 & 12.50 / 26.94 / 8.66 \\ \bottomrule
\end{tabular}
\label{table:c-3-5}
\end{table}

%% file: app-writing/d.tex
We experiment on how compensation methods affect black-box adversarial accuracy, where we craft adversarial examples from a surrogate model without accessing parameters of a target model. Gap between adversarial accuracy against baseline and compensated attacks can indicate whether black-box adversarial accuracy is overestimated originally. We measure black-box transferability of adversarial examples generated using compensation methods among Simple models with different relative width (Table \ref{table:d-1-1}), WRN 28 models with different regularization techniques (Table \ref{table:d-2-1}), and different architectures for CIFAR-10 dataset (Table \ref{table:d-3-1}). Note that we rescale logits ($T=10$) to compensate for the zero loss phenomenon, instead of changing target labels, as mentioned in Section 6.

\subfile{app-writing/table-d-1-1.tex}
\subfile{app-writing/table-d-2-1.tex}
\subfile{app-writing/table-d-3-1.tex}

%% file: app-writing/table-d-1-1.tex
\begin{table}[h]
\centering
\caption{Adversarial accuracy of Simple models with different relative width trained on CIFAR-10 (with fixed weight decay of $5\times10^{-4}$) under the \emph{black-box} setting, where we craft adversarial examples using the source model. We use a PGD attack with $\epsilon=\frac{4}{255}$ in $L_\infty$ norm, and state adversarial accuracy against baseline (before arrow) and targeted (after arrow) attacks. Labels of rows and columns indicate relative width.}
\vspace{0.5em}
\small
\begin{tabular}{|c|c|c|c|c|c|}
\hline \diagbox[innerwidth = 2cm, height = 4ex]{Source}{Target}
   & 1                                                & 2                                                & 4                                                & 8                                                & 16                                               \\ \hline
1  & \cellcolor[HTML]{333333}{\color[HTML]{000000} -} & $51.80 \to 46.05$                                & $61.55 \to 56.30$                                & $64.33 \to 59.29$                                & $62.82 \to 57.11$                                \\ \hline
2  & $40.12 \to 34.67$                                & \cellcolor[HTML]{333333}{\color[HTML]{000000} -} & $48.55 \to 42.28$                                & $50.23 \to 44.17$                                & $50.49 \to 44.26$                                \\ \hline
4  & $39.83 \to 35.55$                                & $37.23 \to 32.51$                                & \cellcolor[HTML]{333333}{\color[HTML]{000000} -} & $38.59 \to 33.48$                                & $39.29 \to 34.42$                                \\ \hline
8  & $38.15 \to 33.40$                                & $33.24 \to 29.04$                                & $32.62 \to 28.33$                                & \cellcolor[HTML]{333333}{\color[HTML]{000000} -} & $28.70 \to 25.26$                                \\ \hline
16 & $32.33 \to 27.36$                                & $30.89 \to 26.55$                                & $30.08 \to 26.26$                                & $25.60 \to 22.53$                                & \cellcolor[HTML]{333333}{\color[HTML]{000000} -} \\ \hline
\end{tabular}
\label{table:d-1-1}
\end{table}

%% file: app-writing/table-d-2-1.tex
\begin{table}[h]
\centering
\caption{Adversarial accuracy of WRN 28 models with different regularization techniques trained on CIFAR-10 under the \emph{black-box} setting, where adversarial examples are crafted using the source model as a surrogate. These models are independently trained under their own training conditions, but are identically initialized. We use the same attack method as in Table \ref{table:d-1-1}}
\vspace{0.5em}
\scriptsize
\begin{tabular}{|c|c|c|c|c|c|c|c|}
\hline \diagbox[innerwidth = 2cm, height = 4ex]{Source}{Target}
      & None                     & Weight decay                       & Weight decay excess                    & Spectral                     & Orthonormal                     & Input gradient                     & Adv train                      \\ \hline
None  & \cellcolor[HTML]{333333} & $22.57 \to 9.54$         & $19.28 \to 7.92$         & $77.51 \to 74.78$        & $28.04 \to 13.99$        & $80.28 \to 75.61$        & $81.34 \to 81.12$        \\ \hline
Weight decay    & $25.37 \to 19.57$        & \cellcolor[HTML]{333333} & $23.72 \to 18.69$        & $79.94 \to 76.40$        & $25.61 \to 19.58$        & $82.73 \to 78.36$        & $81.49 \to 80.87$        \\ \hline
Weight decay excess & $36.27 \to 29.17$        & $39.17 \to 30.34$        & \cellcolor[HTML]{333333} & $81.56 \to 79.29$        & $45.89 \to 35.75$        & $84.66 \to 81.55$        & $81.68 \to 81.08$        \\ \hline
Spectral  & $59.91 \to 56.33$        & $68.59 \to 66.03$        & $65.34 \to 62.75$        & \cellcolor[HTML]{333333} & $70.20 \to 68.18$        & $72.95 \to 70.76$        & $80.40 \to 79.89$        \\ \hline
Orthonormal  & $26.85 \to 12.26$        & $21.04 \to 7.72$         & $26.64 \to 11.86$        & $79.97 \to 77.96$        & \cellcolor[HTML]{333333} & $82.43 \to 79.11$        & $81.53 \to 81.16$        \\ \hline
Input gradient  & $26.21 \to 24.03$        & $33.59 \to 31.31$        & $32.73 \to 30.16$        & $52.61 \to 48.71$        & $36.90 \to 34.40$        & \cellcolor[HTML]{333333} & $79.41 \to 78.47$        \\ \hline
Adv train  & $82.54 \to 81.34$        & $86.25 \to 85.55$        & $83.46 \to 82.40$        & $76.84 \to 75.04$        & $86.07 \to 85.24$        & $80.34 \to 79.53$        & \cellcolor[HTML]{333333} \\ \hline
\end{tabular}
\label{table:d-2-1}
\end{table}

%% file: app-writing/table-d-3-1.tex
\begin{table}[h]
\centering
\caption{Adversarial accuracy of models with different architectures that are trained on CIFAR-10 (with fixed weight decay of $5\times10^{-4}$) under the \emph{black-box setting}. The Simple and Simple-BN models have width 4, and the WRN 28 model has width 2. Details of evaluation is same as in Table \ref{table:d-1-1}.}
\vspace{0.5em}
\small
\begin{tabular}{|c|c|c|c|}
\hline  \diagbox[innerwidth = 2cm, height = 4ex]{Source}{Target}
          & Simple                   & Simple-BN                & WRN 28                   \\ \hline
Simple    & \cellcolor[HTML]{333333} & $51.61 \to 47.81$        & $65.82 \to 62.48$        \\ \hline
Simple-BN & $69.34 \to 66.19$        & \cellcolor[HTML]{333333} & $37.96 \to 31.28$        \\ \hline
WRN 28    & $78.54 \to 74.23$        & $62.15 \to 45.60$        & \cellcolor[HTML]{333333} \\ \hline
\end{tabular}
\label{table:d-3-1}
\end{table}

%% file: app-writing/part-e.tex
Here we elaborate on experimental setup for Section 6. We introduce each model considered, and methods to obtain the lower bound of each model. We follow data preprocessing of \citeauthor{wong2017provable} (\citeyear{wong2017provable}) for MNIST and CIFAR-10 dataset, which additionally includes normalization in case of CIFAR-10; the size of perturbation $\epsilon$ is scaled according to the normalization so that the pixel level perturbation size (which assumes 0-255 RGB image encoded with 8-bit) can be preserved. 

\begin{itemize}
    \item MNIST-A, $\epsilon=0.4$ : this model uses \texttt{`small'} model of \citeauthor{wong2017provable} ($\text{LP}_d\text{-CNN}_A$ of \citeauthor{tjeng2018evaluating}), and is trained with the code publicly available made by \citeauthor{wong2017provable} (\texttt{https://github.com/locuslab/convex\_adversarial}). Training hyperparameters are: \texttt{cascade=1, epochs=200, schedule\_length=20, norm\_type=l1\_median, norm\_eval=l1, starting\_epsilon=0.01, verbose=200, batch\_size=20, test\_batch\_size=10, eps=0.4}. To obtain the lower bound of this model, we use verification method with MILP, with publicly available code provided by \citeauthor{tjeng2018evaluating} (\texttt{https://github.com/vtjeng/MIPVerify.jl}). We measure for untargeted adversarial robustness with $\epsilon=0.4$ in $L_\infty$ norm, and find that MILP can provide \emph{exact} robustness (that is, there is no gap between upper and lower bounds obtained by MILP) for this model.
    
    \item MNIST-B, $\epsilon=0.3$ : this model is \texttt{`large'} model of \citeauthor{wong2017provable} ($\text{LP}_d\text{-CNN}_B$ of \citeauthor{tjeng2018evaluating}), and is directly obtained from repository of \citeauthor{wong2017provable} (\texttt{`mnist\_large\_0\_3.pth'}). The lower bound of \emph{robustness} is directly taken from \citeauthor{tjeng2018evaluating}; because they measure lower and upper bounds of \emph{error}, we take $100\%-\texttt{upper bound of error} \text{ from \citeauthor{tjeng2018evaluating}}$ as the lower bound of robustness.  
    
    \item CIFAR-A, $\epsilon=\frac{2}{255}$ : this model is \texttt{`small'} model for CIFAR-10 of \citeauthor{wong2017provable} ($\text{LP}_d\text{-CNN}_A$ of \citeauthor{tjeng2018evaluating}), and is directly obtained from repository of \citeauthor{wong2017provable} (\texttt{`cifar\_small\_2px.pth'}). The lower bound of robustness is directly taken from \citeauthor{tjeng2018evaluating}, as MNIST-B.
    
    \item CIFAR-B, $\epsilon=\frac{8}{255}$ : this model is \texttt{`ResNet'} model for CIFAR-10 of \citeauthor{wong2017provable} ($\text{LP}_d\text{-RES}$ of \citeauthor{tjeng2018evaluating}), and is directly obtained from repository of \citeauthor{wong2017provable} (\texttt{`cifar\_resnet\_8px.pth'}). The lower bound of robustness is directly taken from \citeauthor{tjeng2018evaluating}, as MNIST-B. 
\end{itemize}

Adversarial accuracy is measured using a PGD attack with stated $\epsilon$ for each model in $L_\infty$ norm. Baseline attacks use 5 random starts to have the same number of evaluations as compensated attacks. 

%% file: app-writing/f.tex
In this section, we analyze how C\&W attack \cite{Carlini_2017} can be affected by the zero loss phenomenon depending on the choice of the objective function. \citeauthor{Carlini_2017} proposed 7 different objective functions that are minimized only when the prediction is wrong. The default choice among them, which gives high success rate with small perturbation size, directly operates on pre-softmax logits $z$:
\begin{gather}
    g_{obj}(x, t) = \max\{-\max\{z_i; i\neq t\}+z_t, -\kappa\}
\end{gather}
where $\kappa$ can act as a confidence parameter. When we use this objective function along with another objective to minimize the perturbation size in $L_2$ norm for C\&W attack, we obtain 100\% success rate for two Simple models (width=4) trained on CIFAR-10 with and without weight decay of $5\times10^{-4}$. The average distortion sizes for two models are 0.188 for the model trained without weight decay (i.e., without any explicit regularization) and 0.214 for the model trained with weight decay.

However, when we choose the objective function that operates on softmaxed logits $\texttt{softmax}(z)$ or cross-entropy loss $l(z, t)$, for example,
\begin{gather}
    g_{obj}(x, t) = 1 - l(f(x), t)
\end{gather}
we observe that the success rate of C\&W attack drops to 88.62\% for the model trained without explicit regularization, and 93.41\% for the model trained with weight decay. Samples on which this C\&W attack fails have zero cross-entropy loss, thus computing gradients for $g_{obj}$ does not produce meaningful optimization direction, similar as in the zero loss phenomenon for bounded first-order attack methods discussed in Section 3.1. Although we do not examine unbounded attack methods, such as C\&W attack discussed here, this analysis shows that unbounded attack methods can suffer from the zero loss phenomenon when the objective function for optimization is not carefully chosen. Objective functions that directly operate on pre-softmax logits, as in the default case for C\&W attack, are preferrable as logits mostly vary linearly, as opposed to loss value that can vary exponentially. 